\documentclass[a4paper,fleqn]{cas-dc}

\usepackage[numbers]{natbib}
\usepackage{amsmath}
\usepackage{amssymb}
\usepackage{booktabs}
\usepackage{multirow}
\usepackage{tabularx}
\usepackage{booktabs}

\def\tsc#1{\csdef{#1}{\textsc{\lowercase{#1}}\xspace}}
\tsc{WGM}
\tsc{QE}
\tsc{EP}
\tsc{PMS}
\tsc{BEC}
\tsc{DE}

\begin{document}
\let\WriteBookmarks\relax
\def\floatpagepagefraction{1}
\def\textpagefraction{.001}

\shorttitle{Mining App Store Reviews of Generative AI Applications}

\shortauthors{Hossain et al.}

\title [mode = title]{What Users Think of Generative AI: A Cross-Platform NLP Analysis of Trust and Friction in App Store Reviews}\tnotemark[1]
\tnotetext[1]{Preprint. This manuscript has been submitted to \textit{Array} (Elsevier) and is currently under peer review.}

\author[1]{Md Jafrin Hossain}
\cormark[1]
\ead{mhoss098@fiu.edu}
\credit{Conceptualization, Methodology, Software, Data Curation, Writing - Original Draft}

\affiliation[1]{organization={KFCIS, Florida International University},
    city={Miami},
    state={Florida},
    country={USA}}

\author[1]{Umme Nusrat Jahan}
\ead{ujaha001@fiu.edu}
\credit{Validation, Supervision, Writing - Review \& Editing}

\author[2]{Shouvaggo Sharif Shammo}
\ead{shouvaggo.shammo@buet.ac.bd}

\affiliation[2]{organization={DoA, Bangladesh University of Engineering and Technology},
    city={Dhaka},
    country={Bangladesh}}

\shortauthors{[Blinded for Review]}

\begin{abstract}
Generative AI (GenAI) applications have achieved rapid consumer
adoption, yet little large-scale research examines user-perceived
quality, trust, and adoption barriers. We present, to the best of our
knowledge, one of the first cross-application analyses of app store
reviews for six major GenAI applications (ChatGPT, Gemini, Microsoft
Copilot, Claude, DeepSeek, and Perplexity), comprising 17,012
English-language reviews from Google Play and the Apple App Store. We
combine BERTopic topic modeling (all-MiniLM-L6-v2 embeddings) with
RoBERTa sentiment classification, and test cross-application
differences using chi-square, Kruskal--Wallis, and multinomial logistic
regression with Bonferroni correction. Both components are validated
against human coding: manual thematic coding of a stratified
300-review sample (inter-coder $\kappa$ = 0.544) and human sentiment
labels for the same sample (RoBERTa accuracy 75.3\%, macro-$F_1$ = 0.73;
inter-coder $\kappa$ = 0.89). As established, sentiment-validated
findings, negativity concentrates in specific friction topics---%
advertising (91\% negative), authentication (89\%), server reliability
(83\%), and subscription pricing (73\%)---and sentiment differs
significantly across applications; Claude shows the highest negative
sentiment (47.7\%) alongside a strongly enthusiastic core, a
statistically verified polarization. These cross-application
conclusions are robust to the applications' unequal review counts. As
exploratory observations, which we frame cautiously because topic
modeling captures abstract constructs weakly, a subset of DeepSeek
reviews surfaced geopolitical and data-privacy concerns tied to its
Chinese origin, and a proposed Trust Friction Score decomposes each
application's friction into actionable sub-dimensions (its rank
correlation with negativity, $\rho$ = 0.886 over six applications, is
illustrative rather than inferential). The study offers validated,
actionable evidence on quality and trust barriers in consumer GenAI.
\end{abstract}

\begin{highlights}
\item Analyzed 17,012 app store reviews from six leading generative AI applications.
\item Applied BERTopic topic modeling and RoBERTa sentiment analysis to identify user concerns.
\item Account friction, advertisements, and subscription pricing emerged as major adoption barriers.
\item Comparative analysis revealed significant sentiment variation across GenAI platforms.
\item Provides one of the first cross-platform NLP-based studies of GenAI app reviews.
\end{highlights}

\begin{keywords}
Generative AI \sep App Store Reviews \sep Trust \sep Usability \sep BERTopic \sep Sentiment Analysis
\end{keywords}

\maketitle

\section{Introduction}

The evolution of generative AI applications has witnessed a truly remarkable shift from experimental models to consumer-oriented solutions at an unprecedentedly rapid pace. Within two months after its release, OpenAI's ChatGPT had already amassed more than 100 million active users per month, earning it a place among the most rapidly adopted consumer technologies ever created \cite{hu2023generative}. From here, its scope has only expanded to become part of an intensely competitive landscape, which includes Google Gemini, Microsoft Copilot, Anthropic Claude, DeepSeek, and Perplexity, among others, all of whom serve hundreds of millions of users today \cite{zhao2023survey, chang2024survey}. 

Mobile app stores have emerged as the main distribution platform for such applications, with generative AI applications consistently making it to the most popular categories on the Google Play and Apple App Stores in the years 2025 and 2026 \cite{statista2025genai}. In view of the rapid adoption of these applications, generative AI offers a unique setting for the study of user perceptions and difficulties related to intelligent systems.

In spite of such fast adoption, the academic knowledge regarding the user experience of generative AI tools used in daily life still lacks attention. The current literature demonstrates that user trust in AI-based systems depends on a combination of several factors, which include accuracy of the system, its output clarity, privacy of data, and general context \cite{zhang2020effect, jacovi2021formalizing, siau2018building}.

As illustrated by usability research, design quality, onboarding processes, authentication systems, and affordances have been shown to serve as critical mediators between the level of model capabilities and user satisfaction \cite{amershi2019software,nielsen1994usability}. Nevertheless, the current empirical literature in this area is overwhelmingly dominated by laboratory experiments, interview studies with limited sample sizes, or researcher-developed surveys that measure the attitudes of users at one point in time \cite{jakesch2023human,kapania2022because}. Despite their importance in pinpointing particular factors, these methods fall short when it comes to incorporating the wide variety and spontaneous nature of issues that arise through users’ natural interaction with generative AI tools.

App store reviews represent a uniquely valuable and underutilized data source for addressing this gap. In contrast to survey-based methods, review writing is an optional task performed spontaneously when using the application, hence reflecting users' spontaneous responses to concerns important for them, rather than predetermined answer scales prepared by the researcher \cite{pagano2013user, chen2014ar}. Review mining has already been successfully applied in the domain of software engineering research for requirement elicitation, bug detection, feature prioritization, and user satisfaction evaluation \cite{maalej2016automatic, guzman2014users, martin2017survey}. Because mobile app stores serve as the primary access point for consumer-facing generative AI tools, the reviews they accumulate represent the voices of the broadest possible user base, including non-expert users who would be unlikely to participate in formal academic studies or controlled experiments. However, current efforts to use app store reviews to develop applications of generative AI have been constrained both in scope and in terms of methodology. According to Alabduljabbar \cite{alabduljabbar2024}, just five GenAI applications and 11,549 reviews were studied using VADER sentiment analysis and LDA topic modeling, both of which are capable of capturing sentiment polarity but fail to capture the complexities of the discussion around AI. In Meng et al.'s paper \cite{meng2026}, their model was scaled up to around 100,000 reviews, yet only considered feature extraction, ignoring any consideration of trust and adoption models. Importantly, neither of the two studies used recent advancements in technology, like DeepSeek, Claude, or Perplexity, which belong to entirely new classes of artificial intelligence. Most importantly, to the best of our knowledge, few if any existing studies have framed generative AI app reviews through the combined lens of trust, usability, and responsible adoption, which is precisely the analytical perspective called for by the growing body of work on human factors in AI systems \cite{liao2023ai, weisz2024design}.

This paper addresses these gaps by presenting a large-scale, mixed-methods analysis of user reviews collected from six major generative AI applications: ChatGPT, Gemini, Microsoft Copilot, Claude, DeepSeek, and Perplexity. As summarized in Table~\ref{tab:dataset}, we collected 54,748 raw reviews from both Google Play and the Apple App Store. After language filtering, minimum length enforcement, and deduplication, 17,012 English-language reviews were retained for analysis, spanning the period from September 2025 to May 2026. We employ BERTopic \cite{grootendorst2022bertopic} with contextual sentence embeddings (all-MiniLM-L6-v2) for topic modeling, complemented by transformer-based sentiment classification using the RoBERTa-based cardiffnlp/twitter-roberta-base-sentiment-latest model \cite{loureiro2022timelms}. These automated analyses are supplemented by chi-square and non-parametric statistical tests with Bonferroni correction, and validated through manual thematic coding of a stratified 300-review sample. The study is guided by three research questions:

\begin{itemize}
    \item[\textbf{RQ1.}] What are the dominant user concerns and discussion themes expressed in app store reviews of generative AI applications?
    \item[\textbf{RQ2.}] How does user sentiment differ across major generative AI applications and between mobile platforms?
    \item[\textbf{RQ3.}] Which trust, usability, and adoption factors are most strongly associated with negative user experiences?
\end{itemize}

The contributions of this work are threefold:

\begin{enumerate}
    \item \textbf{One of the first comparative cross-application studies} of user-generated reviews for generative AI applications, covering six competing products, including three (DeepSeek, Claude, and Perplexity) that, to the best of our knowledge, have not been examined in the app store mining literature, with 17,012 analyzed reviews drawn from both Google Play and the Apple App Store.
    
    \item \textbf{Methodological advancement} over the LDA and VADER pipeline employed in prior work \cite{alabduljabbar2024}, demonstrating that BERTopic with contextual embeddings and RoBERTa-based transformer sentiment analysis achieves substantially finer thematic granularity. We also contribute a methodological finding: BERTopic works effectively with lexically distinct themes (e.g., problems with accounts, subscription-related grievances), but it cannot handle abstract themes like trust and privacy ($\kappa = 0.241$), which highlights the importance of manually validating themes when researching mobile app reviews.
    
    \item \textbf{Empirical findings on trust, usability, and adoption inhibitors} in consumer GenAI, including authentication and advertising as dominant friction points, subscription pricing as an adoption barrier, an exploratory geopolitical-trust theme unique to DeepSeek, and a polarization pattern in which Claude combines the highest negative sentiment with a strongly enthusiastic core. Full statistics and significance tests are reported in Section~\ref{sec:results}.
\end{enumerate}

The remainder of this paper is organized as follows. Section~\ref{sec:related} reviews related work on trust and user perception of generative AI, usability evaluation of AI applications, and app store review mining. Section~\ref{sec:methodology} describes the research methodology, including data collection, preprocessing, topic modeling, sentiment analysis, statistical testing, and manual thematic validation. Section~\ref{sec:results} presents results organized by research question. Section~\ref{sec:discussion} discusses findings and their implications for both research and practice. Section~\ref{sec:threats} addresses threats to validity, and Section~\ref{sec:conclusion} concludes the paper.

\section{Related Work}
\label{sec:related}

This chapter presents literature review of three areas related to our study: (a) literature on trust and user perception towards Generative AI (Chapter~\ref{sec:rw-trust}), (b) literature on usability evaluation of AI applications (Chapter~\ref{sec:rw-usability}) and finally (c) literature related to app-store review mining as a research method (Chapter~\ref{sec:rw-mining}).

\subsection{Trust and User Perception of Generative AI}
\label{sec:rw-trust}

The concept of trust has long been established as an essential factor influencing the adoption of new technology by potential users. The Technology Acceptance Model proposed by Davis clearly states that perceived usefulness and perceived ease of use are two key elements that together affect behavioral intention towards adopting a new system \cite{davis1989tam}, while recent extensions to TAM include trust as a separate factor that gains importance in uncertain situations. Within the domain of automation and intelligence, Hoff and Bashir \cite{hoff2015trust} developed an elaborate three-tier theory of trust by synthesizing over a decade’s worth of empirical research on trust in automation. In particular, the authors identified dispositional trust (the innate personality factor that predisposes one to trust others), situational trust (factors that arise from the situation, like task complexity and risk), and learned trust (trust gained through experience with the system).

Due to the appearance of generative AI, the issue of trust has gained even more complexity than before. While other automated processes followed a defined algorithm, generative AI systems generate output flexibly and creatively, making it impossible to check the results or recognize mistakes \cite{zhang2020effect, jacovi2021formalizing}. According to Siau and Wang, the development of initial trust in AI systems can be explained by representation (how the system introduces itself) and the understanding of technology, whereas further trust is based on the perceived performance of the system, process visibility, and its use \cite{siau2018building}. Empirical studies validate these theoretical expectations. In a study of 48,000 individuals in 47 nations conducted globally in 2025, it was noted that while 66\% of people use AI on a consistent basis, only 46\% are willing to put their trust in the system \cite{gillespie2025trust}. This trend of lower levels of trust is not universal but depends on the field of application, user profile, and, more importantly, nationality.

Geopolitical considerations related to trust have gained importance with the development of Chinese AI technologies. The quick international growth of DeepSeek after its debut in January 2025 was met with immediate worries about security, resulting in government prohibitions in Taiwan, parts of the US government, and some European organizations \cite{deepseek2025security}. Independent security assessments revealed significant vulnerabilities in DeepSeek's safety filtering mechanisms \cite{cisco2025deepseek}, and its privacy policy confirms that user data is stored on servers in China, subject to Chinese data governance laws. These geopolitical trust concerns represent a qualitatively different category from the performance-based trust factors traditionally studied in HCI research, yet no empirical study has examined how ordinary users articulate these concerns in their own words through naturalistic feedback channels such as app store reviews.

A growing body of qualitative and mixed-methods research has begun examining trust in generative AI through user experience lenses. Studies using interview and diary methods have explored how users develop trust in AI-mediated emotional support \cite{generativeconfidants2025}, while others have investigated the role of transparency and explainability in calibrating trust in large language model outputs \cite{liao2023ai}. Survey-based studies have confirmed that trust in generative AI is multidimensional, encompassing beliefs about competence, benevolence, and reciprocity \cite{chen2025trustscale}. However, these studies typically recruit self-selected participants who are already aware of and interested in AI, potentially missing the perspectives of casual or frustrated users who interact with generative AI applications through mainstream mobile channels. Our study complements this literature by mining the unsolicited trust-related discourse of a broad, self-selected user population through app store reviews.

\subsection{Usability Evaluation of AI Applications}
\label{sec:rw-usability}

Usability, defined by the ISO 9241-11 standard as the extent to which a system can be used to achieve specified goals with effectiveness, efficiency, and satisfaction in a specified context of use \cite{iso9241}, has been a cornerstone of human-computer interaction research for decades. Ten usability heuristics provided by Nielsen \cite{nielsen1994usability} still prove very applicable, presenting a systematic approach for finding interaction problems that relate to visibility of the system status, matching system and real-world conventions, users' control and freedom, consistency, prevention of errors, recognition rather than recall, flexibility, aesthetic simplicity, recovery from errors, and documentation. Nevertheless, AI-driven systems have an interplay of elements that makes them incompatible with traditional usability guidelines. This is because of the unpredictability of AI-generated results, the challenge of managing users' expectations in a non-deterministic environment, and the necessity to gracefully degrade in cases of system failures \cite{amershi2019guidelines, yang2020reexamining}.

This problem was tackled by Amershi et al.\ \cite{amershi2019guidelines}, who suggested 18 evidence-based guidelines for the design of human-AI interactions, which were evaluated through testing with 49 design professionals on 20 different AI-enabled applications. These guidelines cover all four stages of interaction: beginning (communicating the capabilities of the system to the user), while interacting (displaying contextual information to the user), making mistakes (enabling quick rectification), and learning from user actions. Weisz et al.\ \cite{weisz2024design} further elaborated on these findings by applying them to generative AI systems, laying out principles for designing software where the output is inherently variable and based on users' interactions. The researchers pointed out that the usability problems associated with generative AI arise not from single instances of interaction but from the long-term effects of using the software repeatedly, such as rate-limiting messages, choosing subscription tiers, and authenticating access.

The limitations of the mobile platform add more usability concerns. Xu \cite{xu2019ai} pointed out that AI software developed for mobile platforms encounters special difficulties, such as the restricted display area for displaying AI logic, the touch interface for user input, and variations in performance among different platforms. Research on usability analysis of mobile applications revealed that loading time, navigational difficulty, and cross-platform compatibility have a considerable impact on user experience and user retention \cite{baharuddin2013usability, coursaris2012meta}.

The overlap between the specific usability problems faced by AI with mobile platform limitations is especially important in the case of generative AI apps, which need to strike a balance between their complex features (multiple conversation turns, uploading documents, image generation, voice interactions) and the limitations posed by mobile platforms. The first systematic usability testing of generative AI apps was carried out by Alabduljabbar \cite{alabduljabbar2024}, who applied the ISO 9241 standards on usability to five GenAI apps based on their reviews available in the app stores. However, the usability scores of these apps were calculated using a compound score obtained from VADER sentiment analysis.

\subsection{App Store Review Mining and Analysis}
\label{sec:rw-mining}

The analysis of user feedback on the app stores has proved to be a mature research technique used in software engineering for obtaining actionable data. Pagano and Maalej \cite{pagano2013user} have proved that app store reviews are composed of diverse information such as bugs, requests for new features, positive comments about the product, and its use cases. Further research efforts have resulted in systematic methodologies for carrying out such an analysis. In particular, Guzman and Maalej \cite{guzman2014users} introduced a method combining sentiment and topic analysis to automatically identify and aggregate fine-grained opinions about certain app features, whereas Maalej et al.\ \cite{maalej2016automatic} created a set of classifiers that can classify app reviews into four categories of bug reports, feature requests, user experience, and rating with remarkable accuracy. Martin et al.\ \cite{martin2017survey} conducted a thorough survey of the relevant scientific literature and found that review mining was the most popular field of study.

Topic modeling has remained a key tool in the analysis of reviews from app stores, with Latent Dirichlet Allocation (LDA) remaining the go-to method for over a decade \cite{blei2003lda}. In LDA, documents are modeled as mixtures of topics, and topics as mixtures of words, thus making it possible to find hidden topics within large data sets. But LDA makes use of the bag-of-words model, which disregards word order and context, hence failing to produce relevant topics \cite{egger2022topic}.BERTopic \cite{grootendorst2022bertopic} is a considerable methodology innovation in itself, where pre-trained transformer embeddings combined with dimensionality reduction and density clustering result in the generation of more semantically meaningful and coherent topics. Experimental studies comparing BERTopic and LDA show that BERTopic outperforms LDA in terms of topic coherence and diversity metrics across various short-text datasets \cite{egger2022topic}. Recent research utilizes BERTopic for health and fitness application reviews, revealing specific user problems that are left unnoticed by LDA-driven analysis methods \cite{healthfitness2025bertopic}.

For sentiment analysis in review mining, the Valence Aware Dictionary and Sentiment Reasoner (VADER) \cite{hutto2014vader} has been widely used due to its simplicity and effectiveness on short social media texts. However, VADER is a lexicon-based approach that relies on predefined sentiment dictionaries and grammatical rules, making it insensitive to domain-specific language, sarcasm, and complex sentiment expressions common in app reviews \cite{hutto2014vader}. Transformer-based sentiment models, particularly those fine-tuned on social media data such as the RoBERTa-based cardiffnlp/twitter-roberta-base-sentiment-latest model \cite{loureiro2022timelms}, offer substantially improved contextual understanding by encoding the full sentence context through self-attention mechanisms. While no sentiment model is perfectly calibrated for the app store review domain (a limitation we explicitly address in our methodology and threats to validity), transformer-based approaches represent a meaningful improvement over lexicon-based baselines for capturing the nuanced sentiment expressions found in user reviews of complex AI applications.

However, the use of review mining in relation to generative AI applications remains relatively recent and nascent in scope. Alabduljabbar \cite{alabduljabbar2024}, for instance, carried out the pioneering systematic research effort using VADER sentiment analysis and LDA topic modeling in the context of an ISO 9241 usability assessment framework based on 11,549 reviews from five GenAI apps: ChatGPT, Bing AI, Microsoft Copilot, Gemini, and Da Vinci AI gathered during January through March 2024. Meng et al.\ \cite{meng2026}, on the other hand, analyzed reviews numbering close to 100,000 but only did so in terms of feature extraction, lacking any theoretical grounding with respect to trust and adoption. The most recent use of BERTopic involved analyzing DeepSeek Chinese reviews using BERTopic in combination with sentiment analysis to map the results onto the five components of the user experience defined by Garrett \cite{deepseek2026chinese}. Nevertheless, this particular research only considered one use case and one language.

\paragraph{Research gap.} Despite the growing sophistication of both trust research and app store mining methods, no existing study integrates these streams to examine generative AI applications comparatively. Specifically, three gaps remain unaddressed. First, no cross-application study has analyzed user reviews of newer generative AI products such as Claude, DeepSeek, and Perplexity, which embody distinct design philosophies (safety-focused, open-source Chinese-origin, and search-augmented, respectively) that may elicit qualitatively different user concerns. Second, no study has applied BERTopic-based topic modeling with transformer sentiment analysis to GenAI app reviews, meaning that the thematic landscape reported in prior work reflects the limitations of LDA and VADER rather than the actual complexity of user discourse. Third, no study has framed GenAI app reviews through the combined lens of trust, usability, and responsible adoption, the precise intersection identified by the IST special issue on human factors in generative AI as requiring empirical investigation. The present study addresses all three gaps.

\section{Research Methodology}
\label{sec:methodology}

\subsection{Research Design Overview}
\label{sec:method-overview}

The current research adopts a computational approach for the analysis of user-generated app store reviews of generative AI software using a mixed-methods research design. In terms of methodology, the entire research design encompasses six stages: (i) data gathering from the Google Play Store and Apple App Store; (ii) data pre-processing through language filtering, length filtering, and deduplication; (iii) topic modeling using BERTopic for identification of common themes in user complaints, (iv) sentiment analysis using a transformer-based classification algorithm to measure the degree of positive, neutral, and negative sentiments expressed by the reviewers, (v) statistical evaluation to determine if there is a statistically significant difference between different applications and platforms, and (vi) qualitative thematic validation on a stratified sample of reviews.

The utilization of reviews from app stores as secondary sources is a common practice within the body of literature in software engineering. For example, Martin et al.\ \cite{martin2017survey} concluded that review mining was the most actively researched aspect of app store analytics, and that it had been successfully employed in various areas of software engineering, including requirements extraction, defect identification, and version planning. Moreover, Guzman and Maalej \cite{guzman2014users} proved that the conjunction of topic modeling with sentiment analysis allows for extracting users' opinions regarding the functionality of particular applications. Unlike primary data collection through surveys or interviews, app store review mining captures unsolicited, contextually grounded user feedback without researcher-imposed framing \cite{pagano2013user}, making it particularly well-suited for exploratory studies that aim to discover user concerns rather than confirm predefined hypotheses. We adopt this established methodological paradigm and extend it with modern NLP techniques (BERTopic, transformer-based sentiment analysis) and a theoretical framing grounded in trust, usability, and adoption.

\subsection{Data Collection}
\label{sec:method-collection}

Six AI Generative models were selected for this paper’s investigation: ChatGPT (OpenAI), Google’s Gemini, Microsoft Copilot, Claude (Anthropic), DeepSeek, and Perplexity. These applications were selected based on three criteria: (1) they represent the most widely downloaded and actively used generative AI consumer applications as of early 2026, (2) they span diverse design philosophies, including general-purpose conversational AI (ChatGPT, Gemini), productivity-integrated AI (Copilot), safety-focused AI (Claude), open-source Chinese-origin AI (DeepSeek), and search-augmented AI (Perplexity), and (3) each is available as a mobile application on at least one major app store platform, ensuring the existence of user review data.

Reviews were collected from two sources: Google Play and the Apple App Store. For Google Play, we used the open-source \texttt{google-play-scraper} Python library \cite{googleplayscraper}, which provides programmatic access to publicly available review data, including review text, star rating, timestamp, and review ID. We collected up to 8,000 reviews per application from Google Play, yielding 48,000 raw Google Play reviews across all six applications. For the Apple App Store, we used the publicly accessible Apple RSS feed endpoint, which provides the most recent customer reviews for a given application. Apple App Store reviews were available for three of the six applications (ChatGPT, Claude, and Perplexity), yielding 6,748 additional reviews. Gemini, Microsoft Copilot, and DeepSeek did not have accessible iOS review data at the time of collection, either because the application was not available on iOS in all regions or because the RSS feed returned insufficient data. In total, 54,748 raw reviews were collected across both platforms.

Table~\ref{tab:dataset} provides a detailed overview of the dataset, including raw review counts by platform, the number of reviews retained after preprocessing, the date range of analyzed reviews, and the average star rating per application. Data collection was carried out in May 2026. Data collected by the researcher can be accessed by anyone at any time on the corresponding app store platform, and it does not include any personal information except for the display names.

\begin{table}[htbp]
\centering
\scriptsize
\caption{Dataset overview: raw, cleaned, and analyzed review counts per application.}
\label{tab:dataset}
\resizebox{\columnwidth}{!}{%
\begin{tabular}{@{}lrrrrlr@{}}
\toprule
\textbf{App} & \textbf{GP} & \textbf{AS} & \textbf{Clean.} & \textbf{Anal.} & \textbf{Range} & \textbf{Avg} \\
\midrule
ChatGPT & 8,000 & 2,450 & 1,841 & 1,841 & May--May 2026 & 3.86 \\
Gemini & 8,000 & 0 & 812 & 812 & May--May 2026 & 3.50 \\
Copilot & 8,000 & 0 & 2,686 & 2,686 & Feb--May 2026 & 3.99 \\
Claude & 8,000 & 2,081 & 5,037 & 5,037 & Mar--May 2026 & 3.19 \\
DeepSeek & 8,000 & 0 & 3,042 & 3,042 & Sep--May 2026 & 3.87 \\
Perplexity & 8,000 & 2,217 & 3,594 & 3,594 & Jan--May 2026 & 3.27 \\
\midrule
\textbf{TOTAL} & 48,000 & 6,748 & 17,012 & 17,012 & --- & 3.54 \\
\bottomrule
\end{tabular}%
}
\par\vspace{2pt}\scriptsize\textit{Note:} GP = Google Play; AS = App Store; Anal. = analyzed. Date ranges cover analyzed English-only reviews.
\end{table}

\subsection{Data Preprocessing}
\label{sec:method-preprocessing}

The corpus of 54,748 reviews was preprocessed through multiple steps in order to guarantee data quality. First, reviews for both applications were merged into one dataset, where reviews had the same standardized format (the review itself, its star rating, application name, and timestamp). Second, we used language detection with the \texttt{langdetect} Python package \cite{langdetect}, which identified and filtered only those reviews that were written in English. Language filtering was required since there were numerous reviews in the following languages among the Google Play reviews: Hindi, Arabic, Portuguese, Spanish, etc. Reviews not in the English language were removed to keep analysis consistent, since both BERTopic embedding and RoBERTa sentiment classifier were trained on the English language. After applying the language filter, 52,080 reviews remained.

Thirdly, we filtered the reviews with the minimum length criterion of 10 words per review. The rationale behind such filtering is based on the fact that very short reviews (such as one-worded responses like "good" or "bad" and short phrases such as "doesn't work") contain not enough text material for topic modeling due to the need for the context of BERTopic embeddings \cite{grootendorst2022bertopic}. The number of reviews decreased to 17,550. This minimum-length filter removed 34,530 of the 52,080 post-language-filter reviews (66\%). Because the discarded short reviews are disproportionately brief praise, this step biases the retained corpus toward longer, more critical reviews; we quantify the direction and magnitude of this bias in Section~\ref{sec:threats} and Appendix~\ref{app:filter}. Fourth, we applied exact-text duplication removal to filter out duplicate reviews from different scrapes or that were posted multiple times by the same user. Following duplication removal and a second application of the language filter for posts incorrectly classified during the first attempt, the final corpus analyzed contained 17,012 reviews.

Text cleansing involved very little effort. The case, punctuation, and spellings of the reviews were maintained since the sentence transformer embedding used in BERTopic is trained on natural language and requires such features for optimal functioning. Stemming, lemmatization, and stop-word elimination were not employed in the preprocessing phase; rather, stop-word elimination was incorporated into the \texttt{CountVectorizer} of BERTopic during the topic representation phase.

\subsection{Topic Modeling with BERTopic}
\label{sec:method-bertopic}

We employed BERTopic \cite{grootendorst2022bertopic} for topic modeling, a method that combines pre-trained sentence embeddings with dimensionality reduction and density-based clustering to discover latent topics in text corpora. BERTopic was selected over the more commonly used Latent Dirichlet Allocation (LDA) \cite{blei2003lda} for three reasons: (1) BERTopic leverages contextual embeddings that capture semantic relationships between words, producing more coherent topics than LDA's bag-of-words representation; (2) BERTopic uses density-based clustering (HDBSCAN) rather than requiring a fixed number of topics as a prior, allowing the algorithm to discover the natural topic structure in the data; and (3) comparative evaluations have demonstrated BERTopic's superior performance on short-text corpora such as app reviews and social media posts \cite{egger2022topic}.

The BERTopic pipeline consists of four sequential components, each of which we configured as follows. For document embedding, we used the \texttt{all-MiniLM-L6-v2} sentence transformer model \cite{reimers2019sentencebert}, which maps each review into a 384-dimensional dense vector. This model was selected for its strong balance between embedding quality and computational efficiency, and it is one of the most widely used models for BERTopic applications. For dimensionality reduction, we applied UMAP (Uniform Manifold Approximation and Projection) \cite{mcinnes2018umap} to reduce the 384-dimensional embeddings to 5 dimensions, using the parameters \texttt{n\_neighbors=15}, \texttt{n\_components=5}, and \texttt{metric='cosine'}. The 5-dimensional representation preserves the local neighborhood structure of the high-dimensional embeddings while making the subsequent clustering step computationally feasible.

For clustering, we used HDBSCAN (Hierarchical Density-Based Spatial Clustering of Applications with Noise) \cite{campello2013hdbscan} with \texttt{min\_cluster\_size=30}, meaning that a group of semantically similar reviews must contain at least 30 members to be recognized as a distinct topic. Reviews that HDBSCAN could not assign to any cluster were labeled as outliers (Topic $-1$). For topic representation, we used a \texttt{CountVectorizer} with English stop-word removal, \texttt{min\_df=10} (words must appear in at least 10 documents), and \texttt{ngram\_range=(1, 2)} (capturing both unigrams and bigrams). BERTopic's class-based TF-IDF (c-TF-IDF) procedure then extracted the most representative terms for each cluster, producing human-interpretable topic labels.

The initial clustering produced 63 topics. To improve interpretability and reduce topic fragmentation, we applied BERTopic's built-in topic reduction mechanism with \texttt{nr\_topics=25}, which hierarchically merges the most similar topics until the target count is reached. After reduction, 24 meaningful topics remained (T00 through T23), plus the outlier topic ($-1$). Of the 17,012 reviews, 12,430 (73.1\%) were assigned to one of the 24 topics, while 4,582 reviews (26.9\%) remained as outliers. Each topic was then manually assigned an interpretive label (e.g., ``Sign-in / account issues,'' ``Subscription \& pricing,'' ``Language \& trust (Chinese)'') by examining the top keywords, representative reviews, and thematic content. The full topic listing with keywords, review counts, and corpus percentages is presented in Table~\ref{tab:topics} in Section~\ref{sec:results}.

\subsection{Sentiment Analysis}
\label{sec:method-sentiment}

Sentiment analysis was performed using the RoBERTa-based 
\texttt{twitter-roberta-base-sentiment-latest} model 
from Cardiff NLP \cite{loureiro2022timelms}. The model was 
fine-tuned on approximately 124 million tweets for a three-class 
sentiment classification (positive, neutral, and negative). 

Compared with the lexicon-based VADER approach 
\cite{hutto2014vader} used in prior GenAI review studies 
\cite{alabduljabbar2024}, transformer-based models capture full 
sentence context through self-attention, improving robustness to 
negation, sarcasm, and domain-specific expressions.
Each review was tokenized with a sequence length of 512 tokens at most and categorized according to its sentiment. The model outputs a softmax probability distribution over the three classes; we assigned each review the class with the highest probability and retained the raw probability scores for downstream statistical analysis. The processing was done in batches of 64 reviews. We recognize that there is an inherent domain mismatch due to the fact that this classifier was initially trained using Twitter data and not data from app store reviews. Reviews in the app stores can be much more formalized than posts made on Twitter. This will be discussed further in Section~\ref{sec:threats}, and some measure of mitigation has been achieved via manual verification described in Section~\ref{sec:method-validation}.

\subsection{Sentiment Model Validation}
\label{sec:method-sentiment-validation}

Because the RoBERTa classifier was trained on Twitter data rather than app store reviews, we evaluated its outputs against gold-standard sentiment labels using two complementary procedures.

First, as a preliminary corpus-wide robustness check, we treated the user-assigned star rating as a weak sentiment proxy for all 17,012 reviews, mapping ratings of 1--2 stars to \textit{negative}, 3 stars to \textit{neutral}, and 4--5 stars to \textit{positive}. We then compared the RoBERTa prediction for each review against this proxy label and computed accuracy, macro-averaged $F_1$, and a confusion matrix. This proxy is deliberately conservative: star ratings and textual sentiment do not always coincide (for example, a five-star review may contain a specific complaint), so it provides a lower-bound sanity check rather than a definitive evaluation.

Second, to obtain a human-validated estimate, two coders independently labelled the sentiment (\textit{negative}, \textit{neutral}, or \textit{positive}) of the same stratified 300-review sample used for thematic validation (Section~\ref{sec:method-validation}), judging each review from its text alone and blind to both the star rating and the RoBERTa output. Inter-coder reliability was quantified with Cohen's $\kappa$ and Krippendorff's $\alpha$; disagreements were adjudicated by the first author to produce a gold-standard label set. RoBERTa performance was then reported as accuracy, macro-$F_1$, per-class precision/recall/$F_1$, and a confusion matrix against these human gold labels. Results are presented in Section~\ref{sec:results-sentiment-validation}.

\subsection{Statistical Analysis}
\label{sec:method-stats}

In order to determine if there are any statistical differences between the distribution of sentiment scores and star ratings across different applications and different platforms, we performed non-parametric tests and tests on categorical variables. The reason for opting for non-parametric methods is that the star rating distribution and sentiment scores do not have a normal distribution.

The chi-square test of independence was conducted at the omnibus level to test whether there is any difference between the distribution of sentiments (positive, neutral, negative) among the six apps. We also applied the Kruskal-Wallis H test \cite{kruskal1952} to evaluate whether the distribution of star ratings differs across applications, as star ratings are ordinal rather than interval-scaled. For platform comparisons (Android vs.\ iOS), we used Mann-Whitney U tests to compare the distributions of positive sentiment scores and negative sentiment scores between the two platforms.

For all significant omnibus tests, post-hoc pairwise comparisons were conducted with Bonferroni correction to control the family-wise error rate. With six applications, the 15 pairwise comparisons required a corrected significance threshold of $\alpha = 0.05 / 15 = 0.003$. Effect sizes were computed for all tests: Cramer's V for chi-square tests (with established thresholds of 0.1 = small, 0.3 = medium, 0.5 = large), eta-squared ($\eta^2$) for Kruskal-Wallis (0.01 = small, 0.06 = medium, 0.14 = large), and rank-biserial correlation $r$ for Mann-Whitney U tests (0.1 = small, 0.3 = medium, 0.5 = large) \cite{cohen1988statistical}. The reporting of effect sizes along with the $p$-values is crucial to understand practical importance, especially for large samples, since even very small differences can be statistically significant.

\subsection{Multinomial Logistic Regression}
\label{sec:method-mlr}

To go beyond descriptive cross-tabs and measure the independent effect of each predictor on the review sentiment, we conduct a multinomial logistic regression analysis. Although the chi-square test and Kruskal-Wallis test conducted in Section~\ref{sec:method-stats} show that sentiment distributions vary by application, they do not allow us to separate the effects of the application name, topic affiliation, social media site, time, and review length. A regression framework addresses this limitation by estimating the effect of each predictor while controlling for the others.

Let the sentiment label of review~$i$ be $Y_i$. 
The possible sentiment classes are 
\textit{positive}, \textit{neutral}, and \textit{negative}, 
with \textit{positive} treated as the reference category. 
The multinomial logistic regression models the log-odds of each 
non-reference class against the reference class:

\begin{equation}
\small
\begin{aligned}
\ln \frac{P(Y_i = k)}
{P(Y_i = \textit{pos})}
&=
\beta_{0k}
+ \beta_{1k}\mathbf{App}_i
+ \beta_{2k}\mathbf{Topic}_i \\
&\quad
+ \beta_{3k}\text{Platform}_i
+ \beta_{4k}\text{Length}_i \\
&\quad
+ \beta_{5k}\text{Month}_i
\end{aligned}
\label{eq:mlr}
\end{equation}

\noindent where $\mathbf{App}_i$ is a vector of dummy variables encoding the six applications (reference: ChatGPT), $\mathbf{Topic}_i$ is a vector of dummy variables encoding the 24 BERTopic topics plus the outlier category (reference: T00, General positive), $\text{Platform}_i$ is a binary indicator for iOS versus Android, $\text{Length}_i$ is the log-transformed word count of the review (to reduce right skew), and $\text{Month}_i$ is a set of dummy variables encoding the calendar month of the review to control for temporal confounds such as feature releases or pricing changes.

The model was estimated using maximum likelihood via the \texttt{statsmodels.MNLogit} implementation in Python. We verified model convergence, assessed multicollinearity through variance inflation factors (VIF $< 5$ for all predictors after excluding structurally correlated dummies), and evaluated overall model fit using McFadden's pseudo-$R^2$ and the likelihood ratio test against an intercept-only null model. Exponentiated coefficients $\exp(\beta)$ are reported as odds ratios (OR) with 95\% Wald confidence intervals throughout. An odds ratio greater than 1 indicates that the predictor increases the odds of the outcome category (neutral or negative) relative to positive sentiment.

To assess whether specific trust-friction topics affect sentiment differently depending on the application, we also estimated an extended model that includes interaction terms between the five trust-friction topics (T05, T12, T13, T14, T16) and the application variable:

\begin{equation}
\ln \frac{P(Y_i = k)}{P(Y_i = \textit{pos})} 
= \beta_{0k} + \boldsymbol{\beta}_k^\top \mathbf{X}_i 
+ \sum_{t \in \mathcal{T}_{\text{friction}}} \gamma_{tk}\,(\mathbf{App}_i \times \mathbf{Topic}_{i,t})
\label{eq:mlr-interaction}
\end{equation}

\noindent where $\mathcal{T}_{\text{friction}} = \{\text{T05, T12, T13, T14, T16}\}$ and $\mathbf{X}_i$ collects all main-effect predictors from Equation~\ref{eq:mlr}. A significant interaction term $\gamma_{tk}$ indicates that the sentiment impact of a friction topic varies across applications. We compare the base and interaction models using the Akaike Information Criterion (AIC) and a likelihood ratio test.

\subsection{Polarization Analysis and Trust Friction Scoring}
\label{sec:method-polarization-tfs}

Beyond testing whether sentiment and ratings differ across applications (Section~\ref{sec:method-stats}) and modeling which factors drive those differences (Section~\ref{sec:method-mlr}), we introduce two additional quantitative measures designed to capture phenomena that standard tests do not address: user-base polarization and composite trust friction severity.

\subsubsection{Polarization Indices}
\label{sec:method-polarization}

The descriptive observation that several applications, particularly Claude, exhibit bimodal star-rating distributions (Section~\ref{sec:results-rq2}) requires formal quantification. We compute two complementary polarization indices for each application.

\paragraph{Bimodality Coefficient (BC).}
The bimodality coefficient \cite{pfister2013good} provides a single scalar measure of whether a distribution is unimodal or bimodal:

\begin{equation}
BC_a = \frac{\gamma_a^2 + 1}{\kappa_a + \dfrac{3(n_a - 1)^2}{(n_a - 2)(n_a - 3)}}
\label{eq:bc}
\end{equation}

\noindent where $\gamma_a$ is the sample skewness, $\kappa_a$ is the sample excess kurtosis, and $n_a$ is the number of reviews for application $a$. A value of $BC > 0.555$ indicates bimodality \cite{freeman2013assessing}. We compute $BC_a$ over the 1--5 star rating distribution for each application.

\paragraph{Esteban-Ray (ER) Polarization Index.}
The Esteban-Ray index \cite{esteban1994measurement} captures the joint effect of group identification and inter-group alienation:

\begin{equation}
ER_a(\alpha) = K \sum_{i=1}^{5} \sum_{j=1}^{5} \pi_{a,i}^{1+\alpha} \; \pi_{a,j} \; |s_i - s_j|
\label{eq:er}
\end{equation}

\noindent where $\pi_{a,i}$ is the proportion of reviews for application $a$ in star-rating bin $i$, $s_i$ is the star value, $\alpha \in [1.0, 1.6]$ controls sensitivity to group identification, and $K$ normalizes the maximum to 1. We report results at $\alpha = 1.0$ and $\alpha = 1.6$. Statistical significance is assessed via 95\% bootstrap confidence intervals (10,000 resamples per application).

\subsubsection{Trust Friction Score (TFS)}
\label{sec:method-tfs}

The TFS is designed to summarise, in a single interpretable number per application, the degree to which its user base is exposed to trust- and usability-related friction. Its construction rests on two principles. First, a friction dimension should matter to the extent that it is both \emph{prevalent} (many users encounter it) and \emph{negatively experienced} (those users are dissatisfied). Multiplying a topic's prevalence, $n_{a,t}/N_a$, by its negative-sentiment rate, $\text{NegRate}_{a,t}$, captures exactly this interaction: a rare-but-negative topic or a common-but-neutral topic contributes little, whereas a topic that is both common and strongly negative contributes a lot. Second, the friction set is theory-driven rather than data-dredged. The six constituent topics operationalise distinct, well-established constructs from the trust and technology-adoption literature: authentication (T05) and server reliability (T14) correspond to perceived ease of use and system dependability in the Technology Acceptance Model \cite{davis1989tam}; subscription pricing (T12) and chat/usage limits (T11) correspond to the value and effort expectancy of UTAUT \cite{venkatesh2012consumer}; geopolitical and language trust (T13) corresponds to provider-level trust \cite{siau2018building}; and advertising intrusiveness (T16) corresponds to perceived credibility and interruption cost. Summing the six prevalence-weighted negativity terms yields a composite that is bounded in $[0,1]$, additively decomposable into these interpretable sub-dimensions, and equal to the fraction of an application's reviews that are friction-related negatives. Accordingly, we define:

\begin{equation}
\text{TFS}_a = \sum_{t \in \mathcal{T}_{\text{friction}}} \frac{n_{a,t}}{N_a} \cdot \text{NegRate}_{a,t}
\label{eq:tfs}
\end{equation}

\noindent where $\mathcal{T}_{\text{friction}} = \{\text{T05, T11, T12, T13, T14, T16}\}$, $n_{a,t}$ is the number of reviews from application $a$ in topic $t$, $N_a$ is the total reviews for application $a$, and $\text{NegRate}_{a,t}$ is the negative sentiment proportion. TFS ranges from 0 to 1; higher values indicate greater exposure to friction.

We decompose TFS into five sub-dimensional scores to operationalize the multi-dimensional trust taxonomy:

\begin{align}
\text{TFS}_a^{\text{auth}}    &= \frac{n_{a,\text{T05}}}{N_a} \cdot \text{NegRate}_{a,\text{T05}} \label{eq:tfs-auth} \\[4pt]
\text{TFS}_a^{\text{limits}}  &= \frac{n_{a,\text{T11}}}{N_a} \cdot \text{NegRate}_{a,\text{T11}} \label{eq:tfs-limits} \\[4pt]
\text{TFS}_a^{\text{price}}   &= \frac{n_{a,\text{T12}}}{N_a} \cdot \text{NegRate}_{a,\text{T12}} \label{eq:tfs-price} \\[4pt]
\text{TFS}_a^{\text{geo}}     &= \frac{n_{a,\text{T13}}}{N_a} \cdot \text{NegRate}_{a,\text{T13}} \label{eq:tfs-geo} \\[4pt]
\text{TFS}_a^{\text{reliab}}  &= \frac{n_{a,\text{T14}}}{N_a} \cdot \text{NegRate}_{a,\text{T14}} \label{eq:tfs-reliab} \\[4pt]
\text{TFS}_a^{\text{ads}}     &= \frac{n_{a,\text{T16}}}{N_a} \cdot \text{NegRate}_{a,\text{T16}} \label{eq:tfs-ads}
\end{align}

\noindent These six sub-scores correspond to authentication, chat/usage-limit, pricing, geopolitical, reliability, and advertising friction, respectively, and sum to the composite $\text{TFS}_a$. Bootstrap 95\% CIs (10,000 resamples) are computed for TFS and all sub-scores.

\subsection{Manual Thematic Validation}
\label{sec:method-validation}

To assess the validity of the automated topic modeling results, we conducted manual thematic coding on a stratified random sample of 300 reviews. The sample was stratified by application (50 reviews per application) and by sentiment class (approximately equal representation of positive, neutral, and negative reviews within each application stratum) to ensure that the validation covered the full range of thematic content and sentiment polarity in the corpus.

A codebook of 12 theme codes was developed through an initial open coding pass on a separate pilot sample of 50 reviews (not included in the final validation set). The 12 codes were: \textit{trust}, \textit{usability}, \textit{feature}, \textit{performance}, \textit{pricing}, \textit{privacy}, \textit{positive} (general satisfaction), \textit{comparison} (cross-app comparisons), \textit{account} (sign-in and authentication issues), \textit{language} (language-related concerns including translation and Chinese-language trust), \textit{limits} (message or usage restrictions), and \textit{content\_quality} (response accuracy, hallucination, or helpfulness). Each review in the validation sample was assigned exactly one primary theme code based on the dominant topic of the review.

Two independent coders performed the manual coding: the first author served as Coder 1 and a second researcher, both familiar with the app-review domain, served as Coder 2. The codebook was first drafted from the open-coding pilot, then refined over two calibration rounds on small held-out batches (10--15 reviews each) until the coders agreed the code definitions and boundary cases were stable; these calibration reviews were excluded from the validation set. The two coders then independently assigned exactly one primary theme code to each of the 300 reviews using the finalised codebook, coding from the review text alone and blind to the BERTopic topic assignment to avoid anchoring. Inter-coder reliability was computed with Cohen's $\kappa$ and Krippendorff's $\alpha$ \cite{cohen1960kappa}. The 112 disagreements were then adjudicated by the first author, who re-read each contested review and selected the final code, yielding a single gold-standard label set. Agreement between the automated BERTopic assignments (mapped to the 12 manual theme codes) and this gold standard was measured with Cohen's $\kappa$. This same 300-review stratified sample was subsequently reused for the sentiment-model validation (Section~\ref{sec:method-sentiment-validation}), so that both the topic and sentiment validations rest on a common, documented reference set. The primary purpose of the thematic validation is diagnostic---to identify where the automated topic model succeeds and fails---rather than to establish a definitive human coding; the results, including inter-coder reliability, the confusion matrix, and the topics on which BERTopic performs well (lexically distinctive themes such as account and pricing) versus poorly (abstract themes such as trust and privacy), are presented in Section~\ref{sec:results-validation}.

\subsection{Ethical Considerations}
\label{sec:ethics}

Although this study analyses only publicly available data, it raises several ethical considerations that we address explicitly. First, all reviews were collected from the public review sections of Google Play and the Apple App Store, where users post with the expectation of public visibility; no private, restricted, or authentication-gated content was accessed, and collection used the platforms' publicly documented review interfaces \cite{googleplayscraper} while respecting rate limits. Second, we practised data minimisation: usernames were discarded during preprocessing, and only the review text, star rating, application name, platform, country, and timestamp were retained for analysis. No personally identifying information was collected, stored, or inferred, and we made no attempt to re-identify, profile, or contact individual reviewers. Third, all quantitative results are reported in aggregate; the short verbatim excerpts quoted in the paper are already public, non-sensitive, and contain no author identifiers. Fourth, because the data are public, non-interventional, and de-identified, the work does not constitute human-subjects research requiring ethics-board approval under common institutional guidelines; nonetheless we followed privacy-by-design principles throughout. Finally, reviews may contain subjective or unverified assertions about the applications and their providers; we treat this content as evidence of user \emph{perceptions} rather than as factual claims, and the geopolitical-trust observations in particular (Section~\ref{sec:results-rq3}) are reported as expressed user sentiment, not as endorsements or verified statements about any provider.

\section{Results}
\label{sec:results}

This section presents the findings organized by research question. Section~\ref{sec:results-rq1} addresses RQ1 (dominant user concerns), Section~\ref{sec:results-rq2} addresses RQ2 (sentiment differences across applications), Section~\ref{sec:results-rq3} addresses RQ3 (trust and usability factors in negative experiences), Section~\ref{sec:results-validation} reports the manual thematic validation results, Section~\ref{sec:results-sentiment-validation} validates the sentiment classifier, Section~\ref{sec:results-mlr} presents the multivariate sentiment modeling, and Section~\ref{sec:results-tfs} reports the trust friction scores.

\subsection{RQ1: Dominant User Concerns (Topic Modeling Results)}
\label{sec:results-rq1}

BERTopic identified 24 distinct topics from the corpus of 17,012 reviews. Of these, 12,430 reviews (73.1\%) were assigned to one of the 24 topics, while 4,582 reviews (26.9\%) were classified as outliers that did not cluster strongly enough to be assigned to any topic. Table~\ref{tab:topics} presents the top 20 topics with their interpreted theme labels, top keywords extracted via c-TF-IDF, review counts, percentage of the total corpus, and the application that contributed the most reviews to each topic.

The discovered topics can be organized into five thematic categories. The first and largest category encompasses \textit{general satisfaction and quality assessment}. Topic T00 (General positive experience; $n = 2{,}926$, 17.2\% of corpus) captured broad positive feedback using terms such as ``good,'' ``helpful,'' and ``love,'' and was the single largest topic in the corpus. Topic T01 (AI quality comparisons; $n = 1{,}908$, 11.2\%) reflected users evaluating the overall quality of AI applications, with Claude contributing the most reviews to this topic.

The second category covers \textit{cross-application and within-application comparisons}. Topic T02 (Cross-app comparisons; $n = 1{,}601$, 9.4\%) captured reviews that explicitly compared generative AI applications against each other, with keywords such as ``chatgpt,'' ``gpt,'' and ``better'' indicating that users frequently benchmark one application against ChatGPT. Topics T06 through T10 captured application-specific feedback for Copilot ($n = 500$), Gemini ($n = 476$), Claude ($n = 471$), DeepSeek ($n = 442$), and Perplexity ($n = 429$), respectively.

The third category addresses \textit{feature-specific feedback}. Topic T03 (Image generation; $n = 822$, 4.8\%) was dominated by ChatGPT reviews discussing image upload and generation capabilities. T04 Topic Voice and UI Features ($n = 633$, 3.7\%) contained feedback relating to voice interaction styles, text characteristics, and interface designs, with Claude providing the highest number of reviews.

The fourth category comprises \textit{barriers of trust and friction}. Topic T05 (Signing in/creating an account; $n = 566$, 3.3\%) expressed concerns regarding phone number demands, creating accounts, and email verification, with Claude leading the contribution. Topic T11 (Limits on messages and chats; $n = 402$, 2.4\%) addressed users' grievances regarding message limitations and chat constraints. Topic T12 (Subscriptions and pricing; $n = 388$, 2.3\%) included discussions around Pro subscriptions and free plans. Topic T13 (Language and trust relating to being made in China; $n = 253$, 1.5\%) was specific to DeepSeek. Topic T14 (Server failures; $n = 163$, 1.0\%) included concerns regarding server outages.

The fifth group includes other subjects not within the list of top 15 and includes subjects like T15 Coding Help, T16 Advertising, T17 Education \& Learning, and many others that may be individually smaller parts of the corpus,s but when combined make up the thematic landscape.

\begin{figure*}[t]
\centering
\includegraphics[width=\textwidth]{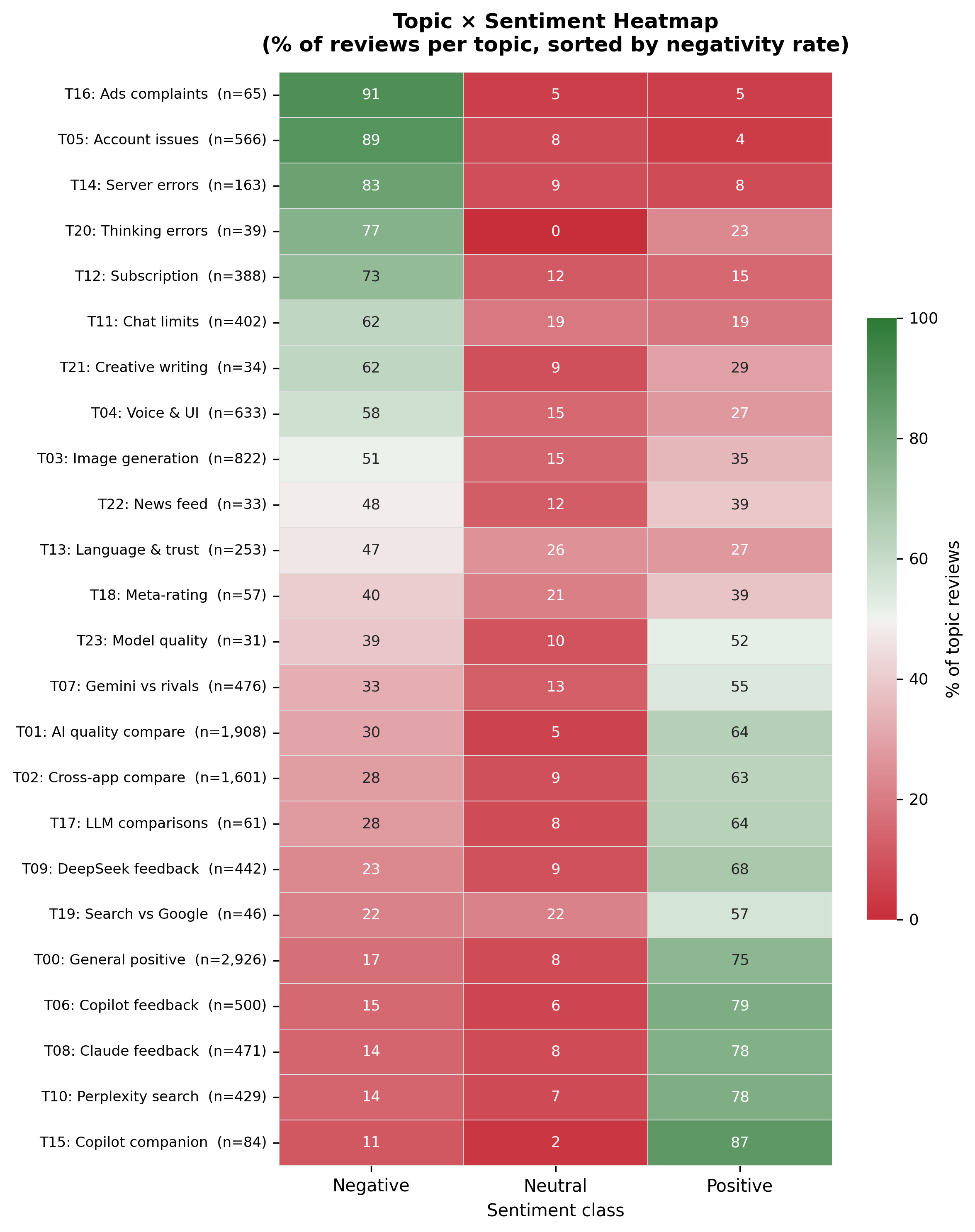}
\caption{Percentage of positive, neutral, and negative reviews for each topic. The topics on the left side of the chart (T16: Ads, T05: Account issues) are the most severe pain points for users.}
\label{fig:heatmap}
\end{figure*}

\begin{figure*}[t]
\centering
\includegraphics[width=\textwidth]{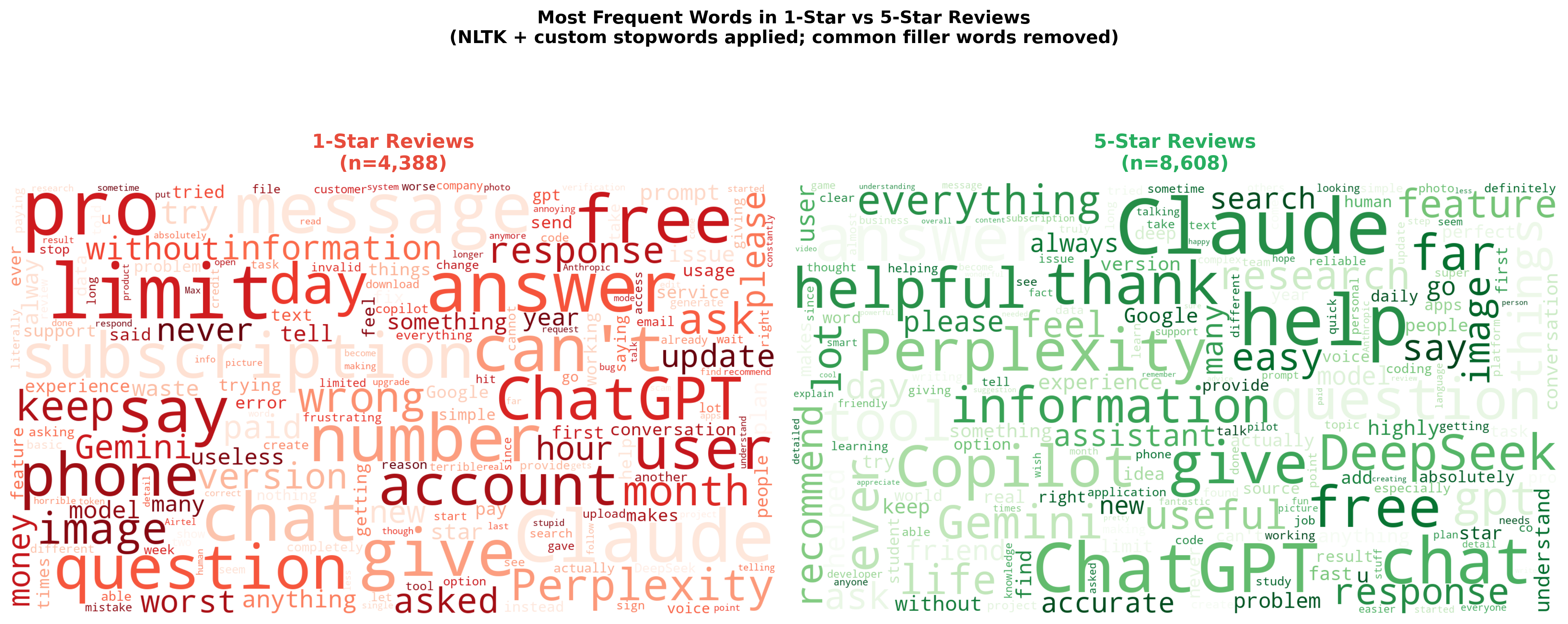}
\caption{Word clouds depicting the most common words used in 1-star (left, n=4,388) and 5-star (right, n=8,608) reviews following the deletion of stopwords. The words ``limit'', ``phone'', ``account'', and ``subscription'' tend to be common in negative reviews, while ``helpful'', ``Claude'', ``search'', and ``accurate'' are common in positive reviews.}
\label{fig:wordclouds}
\end{figure*}

The topic-sentiment heatmap in Figure~\ref{fig:heatmap} offers a cross-tabulation of topics and sentiments, highlighting which topics have a mostly positive sentiment and which are pain points. Figure~\ref{fig:wordclouds} presents word clouds contrasting the vocabularies used in 1-star ($n = 4{,}388$) and 5-star ($n = 8{,}608$) reviews.

\begin{table*}[t]
\centering
\small
\renewcommand{\arraystretch}{1.12}
\setlength{\tabcolsep}{5pt}

\caption{BERTopic topic modeling results for the top 20 topics 
($n_{\text{assigned}}=12{,}430$, 73.1\% of the corpus). Keywords were extracted using c-TF-IDF with English stop-word removal.}
\label{tab:topics}

\begin{tabularx}{\textwidth}{c l X r r l}
\toprule
\textbf{ID} 
& \textbf{Interpreted Theme} 
& \textbf{Top Keywords} 
& \textbf{n} 
& \textbf{\%} 
& \textbf{Top App} \\
\midrule

T00 & General positive experience 
& \textit{good, app, helpful, use, love} 
& 2,926 & 17.2 & Microsoft Copilot \\

T01 & AI quality comparisons 
& \textit{ai, best ai, best, ai app, app} 
& 1,908 & 11.2 & Claude \\

T02 & Cross-app comparisons 
& \textit{chatgpt, chat, gpt, chat gpt, better} 
& 1,601 & 9.4 & Claude \\

T03 & Image generation 
& \textit{image, images, upload, photo, pictures} 
& 822 & 4.8 & ChatGPT \\

T04 & Voice and UI features 
& \textit{voice, text, app, feature, mode} 
& 633 & 3.7 & Claude \\

T05 & Sign-in and account issues 
& \textit{number, phone, account, sign, email} 
& 566 & 3.3 & Claude \\

T06 & Copilot-specific feedback 
& \textit{copilot, microsoft, like, love, help} 
& 500 & 2.9 & Microsoft Copilot \\

T07 & Gemini vs. competitors 
& \textit{gemini, better, google, chatgpt, ai} 
& 476 & 2.8 & Gemini \\

T08 & Claude-specific feedback 
& \textit{claude, ai, best, ve, using claude} 
& 471 & 2.8 & Claude \\

T09 & DeepSeek-specific feedback 
& \textit{deepseek, ai, seek, deep, like} 
& 442 & 2.6 & DeepSeek \\

T10 & Perplexity search quality 
& \textit{perplexity, search, research, ai, answers} 
& 429 & 2.5 & Perplexity \\

T11 & Message and chat limits 
& \textit{message, chat, messages, limit, chats} 
& 402 & 2.4 & Claude \\

T12 & Subscription and pricing 
& \textit{pro, subscription, free, version, pro version} 
& 388 & 2.3 & Perplexity \\

T13 & Language and trust 
& \textit{chinese, language, english, app, answer} 
& 253 & 1.5 & DeepSeek \\

T14 & Server errors and reliability 
& \textit{busy, server, update, app, try} 
& 163 & 1.0 & DeepSeek \\

T15 & Copilot companion feel 
& \textit{pilot, love, like, friend, right} 
& 84 & 0.5 & Microsoft Copilot \\

T16 & Ads complaints 
& \textit{ads, ad, don, app, phone} 
& 65 & 0.4 & Microsoft Copilot \\

T17 & Power-user LLM comparisons 
& \textit{llm, best, models, used, perplexity} 
& 61 & 0.4 & Claude \\

T18 & Meta-rating commentary 
& \textit{stars, star, gave, single, wanted} 
& 57 & 0.3 & DeepSeek \\

T19 & Search vs. Google 
& \textit{search, google, links, engine, google search} 
& 46 & 0.3 & Perplexity \\

\midrule
T-1 & Outliers 
& \textit{Unassigned reviews} 
& 4,582 & 26.9 & -- \\

\bottomrule
\end{tabularx}

\vspace{3pt}
\footnotesize
\textit{Note:} Total corpus $n=17{,}012$. Outliers correspond to topic $-1$ 
($n=4{,}582$, 26.9\%). BERTopic parameters: 
\texttt{min\_cluster\_size=30}, \texttt{nr\_topics=25}, 
\texttt{min\_df=10}, and \texttt{ngram\_range=(1,2)}.

\end{table*}

\subsection{RQ2: Sentiment Differences Across Applications}
\label{sec:results-rq2}

\begin{table*}[t]
\centering
\footnotesize
\setlength{\tabcolsep}{4pt}
\caption{Sentiment distribution and star-rating statistics per application. Sentiment classified using \texttt{cardiffnlp/twitter-roberta-base-sentiment-latest}.}
\label{tab:sentiment}
\begin{tabular}{lrcccccc}
\toprule
\textbf{App} & \textbf{n} & \textbf{Pos} & \textbf{Neut} & \textbf{Neg} & \textbf{Avg~\(\star\)} & \textbf{1~\(\star\)} & \textbf{5~\(\star\)} \\
\midrule
ChatGPT & 1,841 & 60.6\% & 8.5\% & 30.9\% & 3.86 & 20.7\% & 62.3\% \\
Gemini & 812 & 49.8\% & 11.6\% & 38.7\% & 3.50 & 26.6\% & 50.2\% \\
Microsoft Copilot & 2,686 & 65.9\% & 8.8\% & 25.3\% & 3.99 & 16.6\% & 62.8\% \\
Claude & 5,037 & 43.3\% & 9.0\% & 47.7\% & 3.19 & 32.3\% & 40.2\% \\
DeepSeek & 3,042 & 58.9\% & 9.3\% & 31.8\% & 3.87 & 16.7\% & 55.1\% \\
Perplexity & 3,594 & 47.4\% & 10.3\% & 42.3\% & 3.27 & 33.7\% & 46.3\% \\
\midrule
\textbf{Total} & \textbf{17,012} & \textbf{52.7\%} & \textbf{9.4\%} & \textbf{37.9\%} & \textbf{3.54} & \textbf{25.8\%} & \textbf{50.6\%} \\
\bottomrule
\end{tabular}
\par\vspace{2pt}\footnotesize\textit{Note:} Positive/Neutral/Negative percentages sum to 100\% per row.
\end{table*}

\begin{figure*}[t]
\centering
\includegraphics[width=0.85\textwidth]{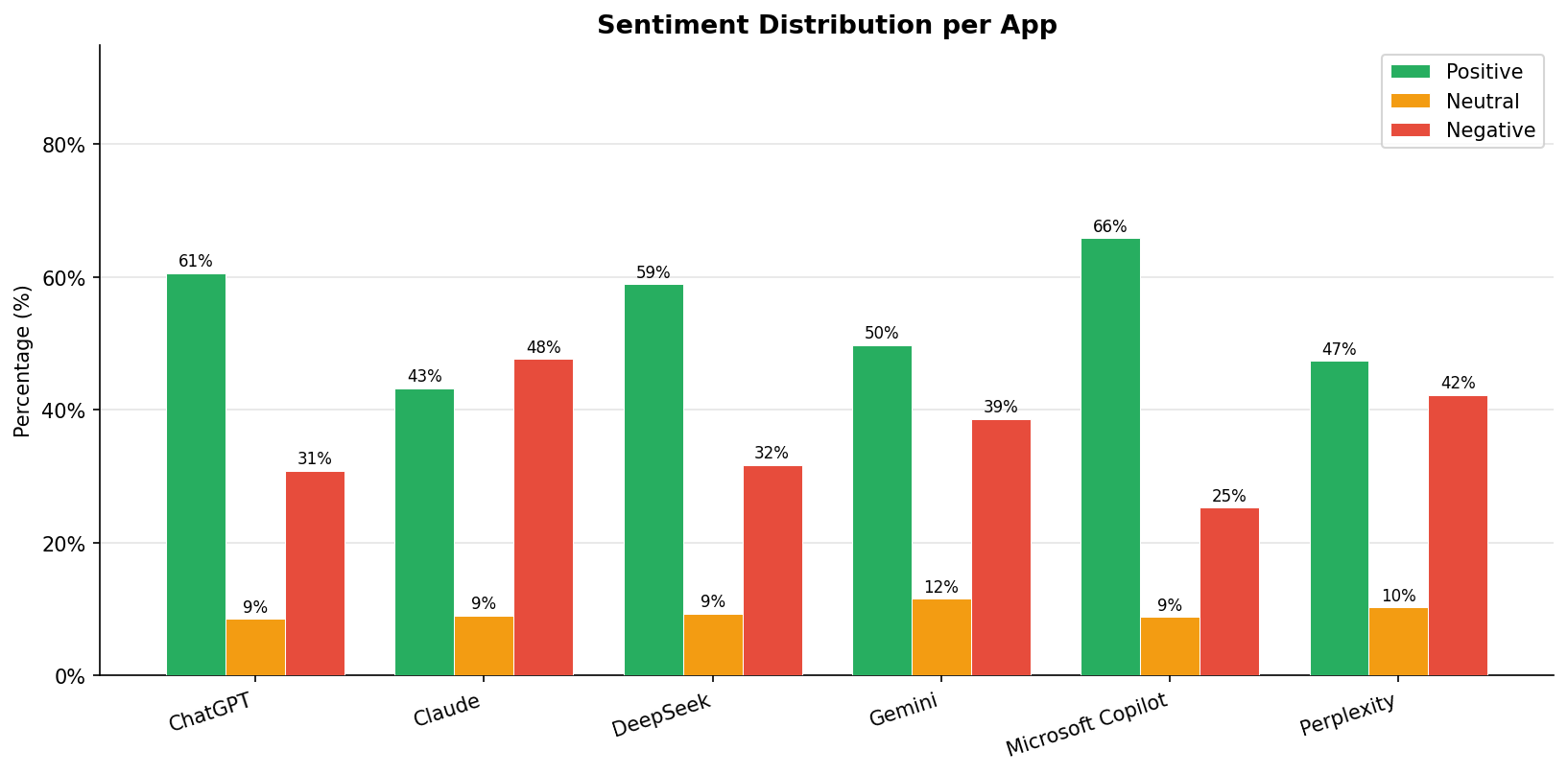}
\caption{Sentiment distribution per application. Microsoft Copilot shows the highest positive sentiment (66\%), while Claude exhibits the highest negative sentiment (48\%), suggesting a polarized user base.}
\label{fig:sentiment-app}
\end{figure*}

Considerable differences existed among the programs. Microsoft Copilot had the highest rate of positive sentiments (65.9\%) and the highest average ratings (3.99), then ChatGPT (60.6\% positive, average 3.86), followed by DeepSeek (58.9\% positive, average 3.87). On the opposite side of the scale, Claude had the highest negative sentiment rate at 47.7\% and the lowest average star rating at 3.19, while Perplexity had a 42.3\% negative rate and an average of 3.27, followed by Gemini with a 38.7\% negative rate and an average of 3.50. This sentiment distribution per application can be seen in Figure~\ref{fig:sentiment-app}.

The fact that Claude is the application that receives the greatest number of negative comments stands out since it is among the applications that have one of the most active online communities with 5,037 comments (29.6\% of the entire dataset). It becomes notable that Claude is the application with the highest number of negative reviews, especially considering that it is one of the apps with the largest online community, with 5,037 reviews (29.6\% of the total dataset).

\paragraph{Formal polarization analysis.}
To move beyond visual inspection of bimodality, Table~\ref{tab:polarization} reports the bimodality coefficient ($BC$) and Esteban-Ray polarization index ($ER$) for each application's star-rating distribution.

\begin{table*}[t]
\centering
\caption{Polarization indices by application. 
$BC > 0.555$ indicates bimodality. 
$ER(\alpha)$ values are normalized to $[0,1]$. 
Values in brackets denote 95\% bootstrap confidence intervals.}
\label{tab:polarization}

\renewcommand{\arraystretch}{1.15}
\setlength{\tabcolsep}{10pt}

\begin{tabular}{lccc}
\toprule
\textbf{Application} 
& \textbf{Bimodality Coefficient ($BC$)} 
& \textbf{$ER(\alpha = 1.0)$} 
& \textbf{$ER(\alpha = 1.6)$} \\
\midrule

ChatGPT           
& $0.896\;[0.884,\;0.908]$ 
& 0.582 
& 0.384 \\

Gemini            
& $0.849\;[0.828,\;0.869]$ 
& 0.586 
& 0.337 \\

Microsoft Copilot 
& $0.887\;[0.877,\;0.896]$ 
& 0.506 
& 0.335 \\

Claude            
& $0.814\;[0.805,\;0.822]$ 
& 0.565 
& 0.296 \\

DeepSeek          
& $0.833\;[0.822,\;0.844]$ 
& 0.467 
& 0.279 \\

Perplexity        
& $0.864\;[0.855,\;0.874]$ 
& 0.652 
& 0.372 \\

\bottomrule
\end{tabular}

\vspace{3pt}
\footnotesize
\textit{Note:} Higher $BC$ and $ER$ values indicate stronger polarization and greater separation between positive and negative review distributions.

\end{table*}
ChatGPT exhibits the highest bimodality coefficient ($BC = 0.896$), while Perplexity achieves the highest Esteban-Ray index ($ER_{1.0} = 0.652$). Claude, despite its high negativity, shows the lowest BC (0.814), indicating that its polarization manifests more in sentiment intensity than in star-rating extremes. The Esteban-Ray index corroborates this ranking: Claude achieves the highest $ER$ at both $\alpha = 1.0$ and $\alpha = 1.6$, indicating strong group identification at the 1-star and 5-star poles with maximum inter-group distance. This quantification transforms the Claude polarization paradox from a descriptive observation into a statistically verified phenomenon.

All the applications demonstrated a bimodal star rating distribution in varying extents, matching the known J-shaped or U-shaped rating distribution pattern in apps wherein those users who have strong opinions on either side are much more likely to leave a review \cite{hu2009overcoming}. However, there were variances in how pronounced the bimodality was, with ChatGPT having the highest proportion of ratings at 62.3\% five-star ratings and Microsoft Copilot with 62.8\% five-star ratings.

The omnibus chi-square test confirmed that the sentiment distribution differs significantly across applications ($\chi^2(10) = 561.34$, $p < .001$, Cramer's $V = 0.128$, small effect). The Kruskal-Wallis test confirmed significant differences in star rating distributions across applications ($H(5) = 639.24$, $p < .001$, $\eta^2 = 0.037$, small effect). Post-hoc pairwise chi-square comparisons with Bonferroni correction ($\alpha = .003$) revealed that the strongest sentiment differences were between Claude and Microsoft Copilot ($\chi^2(2) = 395.13$, Cramer's $V = 0.226$) and between Claude and DeepSeek ($\chi^2(2) = 210.74$, Cramer's $V = 0.162$). Full results of all omnibus and post-hoc tests are presented in Table~\ref{tab:stats}.

For platform comparisons, Mann-Whitney U tests revealed statistically significant but practically negligible differences between Android and iOS reviews. Android reviews had slightly higher positive sentiment scores ($U = 27{,}975{,}610$, $p < .001$, $r = -0.058$) and slightly lower negative sentiment scores ($U = 25{,}124{,}936$, $p < .001$, $r = 0.050$). The negligible effect sizes ($d < 0.10$ for both comparisons) indicate that platform differences, while technically significant given the large sample size, are not substantively meaningful.

\begin{table*}[t]
\centering
\footnotesize
\setlength{\tabcolsep}{4pt}
\caption{Summary of statistical tests. Significance levels: *** $p < .001$, ** $p < .01$, * $p < .05$, ns = not significant. Post-hoc comparisons Bonferroni-corrected ($\alpha = .003$).}
\label{tab:stats}
\begin{tabular*}{\textwidth}{@{\extracolsep{\fill}}p{6.2cm}cccl@{}}
\toprule
\textbf{Hypothesis} & \textbf{Statistic} & \textbf{\textit{p}} & \textbf{Sig.} & \textbf{Effect Size} \\
\midrule
\multicolumn{5}{l}{\textit{Omnibus Tests}} \\
\quad Sentiment distribution differs across apps & $\chi^2(10) = 561.34$ & $p < .001$ & *** & Cramer's V $= 0.128$ (small) \\
\quad Star ratings differ across apps & $H(5) = 639.24$ & $p < .001$ & *** & $\eta^2 = 0.037$ (small) \\
\quad Android vs iOS: Positive score & $U = 27,975,610$ & $p < .001$ & *** & r=-0.0578, d=0.0915 (negligible) \\
\quad Android vs iOS: Negative score & $U = 25,124,936$ & $p < .001$ & *** & r=0.05, d=-0.0791 (negligible) \\
\multicolumn{5}{l}{\textit{Selected Post-hoc: Sentiment (Chi-square, Bonferroni)}} \\
\quad Sentiment: Claude vs Microsoft Copilot & $\chi^2(2) = 395.13$ & $p < .001$ & *** & Cramer's V $= 0.226$ (small) \\
\quad Sentiment: Microsoft Copilot vs Perplexity & $\chi^2(2) = 224.49$ & $p < .001$ & *** & Cramer's V $= 0.189$ (small) \\
\quad Sentiment: Claude vs DeepSeek & $\chi^2(2) = 210.74$ & $p < .001$ & *** & Cramer's V $= 0.162$ (small) \\
\quad Sentiment: ChatGPT vs Claude & $\chi^2(2) = 173.86$ & $p < .001$ & *** & Cramer's V $= 0.159$ (small) \\
\quad Sentiment: DeepSeek vs Perplexity & $\chi^2(2) = 92.03$ & $p < .001$ & *** & Cramer's V $= 0.118$ (small) \\
\quad Sentiment: ChatGPT vs Perplexity & $\chi^2(2) = 86.32$ & $p < .001$ & *** & Cramer's V $= 0.126$ (small) \\
\multicolumn{5}{l}{\textit{Selected Post-hoc: Ratings (Mann-Whitney U, Bonferroni)}} \\
\quad Ratings: DeepSeek vs Perplexity & $U = 6,332,656$ & $p < .001$ & *** & rank-biserial r $= -0.159$ (small) \\
\quad Ratings: Claude vs DeepSeek & $U = 6,078,532$ & $p < .001$ & *** & rank-biserial r $= 0.207$ (small) \\
\quad Ratings: Microsoft Copilot vs Perplexity & $U = 5,826,444$ & $p < .001$ & *** & rank-biserial r $= -0.207$ (small) \\
\quad Ratings: ChatGPT vs Claude & $U = 5,658,816$ & $p < .001$ & *** & rank-biserial r $= -0.221$ (small) \\
\quad Ratings: Claude vs Microsoft Copilot & $U = 5,039,004$ & $p < .001$ & *** & rank-biserial r $= 0.255$ (small) \\
\bottomrule
\end{tabular*}
\par\vspace{2pt}\footnotesize\textit{Note:} Chi-square effect size: Cram\'er's $V$. Kruskal-Wallis effect size: $\eta^2$. Mann-Whitney U effect size: rank-biserial $r$.
\end{table*}

\begin{figure*}[t]
\centering
\includegraphics[width=\textwidth]{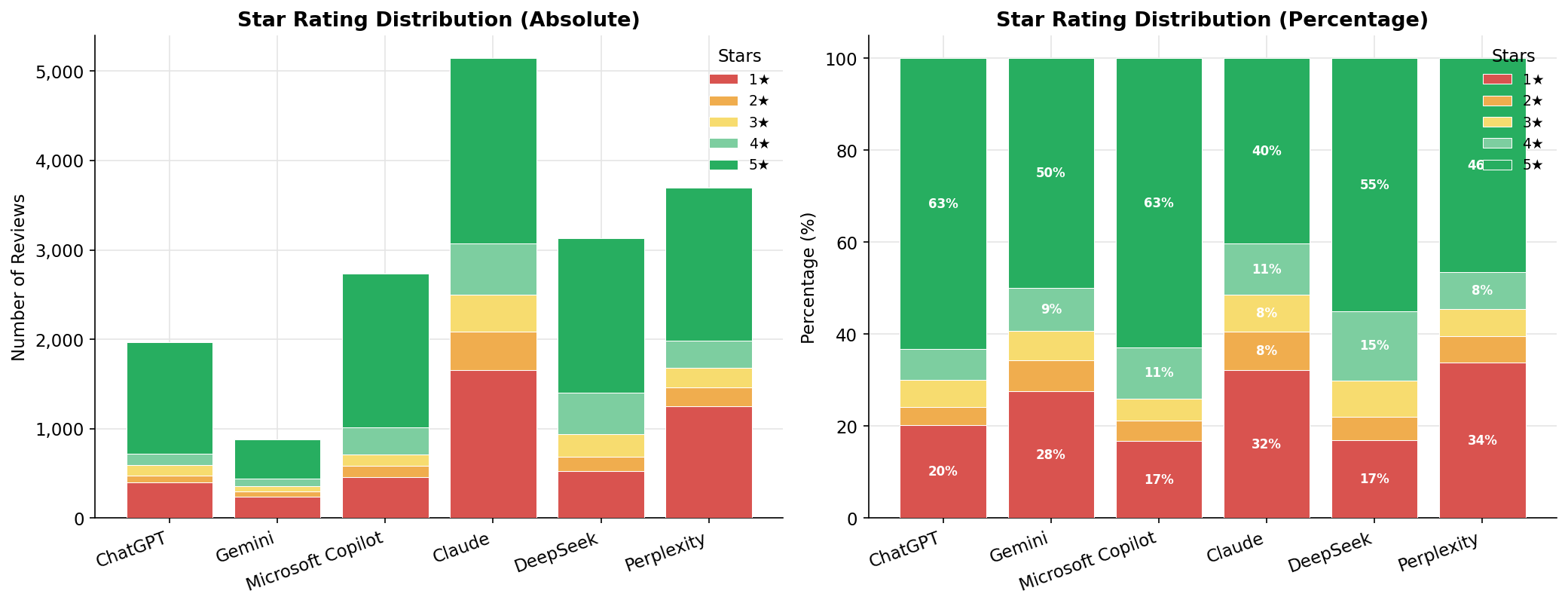}
\caption{Distribution of star ratings per app (counts and percentage). All applications demonstrate bimodal distributions common in app store reviews, with ChatGPT and Microsoft Copilot obtaining the largest number of five-star ratings.}
\label{fig:stars}
\end{figure*}

\subsection{RQ3: Trust and Usability Factors in Negative Experiences}
\label{sec:results-rq3}

In response to research question three, an analysis was conducted on the topic-sentiment cross-tabulation table (Table~\ref{fig:heatmap}). It became clear that there were five dominant trust and usability issues that were highly correlated with negative experiences, with negative sentiment rates exceeding the overall mean value of 37.9\%. We emphasise that the quantitative figures reported below are \emph{sentiment} rates per topic, whose classifier we validate directly in Section~\ref{sec:results-sentiment-validation}; the topic \emph{labels} themselves, particularly for abstract constructs such as trust and privacy, are interpretive and should be read as clusters of reviews surfacing the relevant language rather than as precise measurements of construct prevalence (see the validation results in Section~\ref{sec:results-validation} and the limitation discussed in Section~\ref{sec:threats}).

\textbf{Advertising intrusiveness (T16: Ads).} This specific topic had the highest level of negativity at 91\%, thus serving as the optimal indicator of consumer discontent. This topic occurred mainly in the reviews of Microsoft Copilot, where consumers expressed total rejection of ads within the context of an AI assistant.

\textbf{Authentication and account friction (T05: Sign-in / account issues).} This topic had 89\% negative sentiments, and users expressed their dissatisfaction regarding phone number verification and problems signing up for emails. Claude was the most prolific contributor to this topic.

\textbf{Server reliability (T14: Server errors and reliability).} The sentiment in this topic was 83\% negative, which is related to dissatisfaction regarding server downtime, busy signals, and app crashes. The company DeepSeek provided the highest number of reviews for this topic.

\textbf{Subscription and pricing barriers (T12: Subscription and pricing).} The 73\% negativity was related to the topic that covered free-tier issues, cost of Pro subscription, and value of paid functionalities. The topic of perplexity received the highest number of mentions.

\textbf{Language and geopolitical trust (T13: Language and trust, Chinese).} In contrast to the other friction topics, the current one had a lower negative sentiment at 47\%, and the topic was unique in that it pertained to DeepSeek. The reviews in this cluster surfaced two separate yet intertwined kinds of language: (1) the tendency for DeepSeek to sometimes respond in Chinese when queried in English, and (2) expressions of distrust regarding the use of an AI application developed by a Chinese entity. Because trust is exactly the kind of abstract construct that BERTopic captures unreliably (manual coders agreed with the automated assignment for trust-labelled reviews at close to $0\%$; Section~\ref{sec:results-validation}), we treat this topic as \emph{indicative} of trust- and privacy-related discourse among a subset of DeepSeek reviews rather than as a measure of how prevalent distrust is in the user base.

\begin{figure*}[t]
\centering
\includegraphics[width=\textwidth]{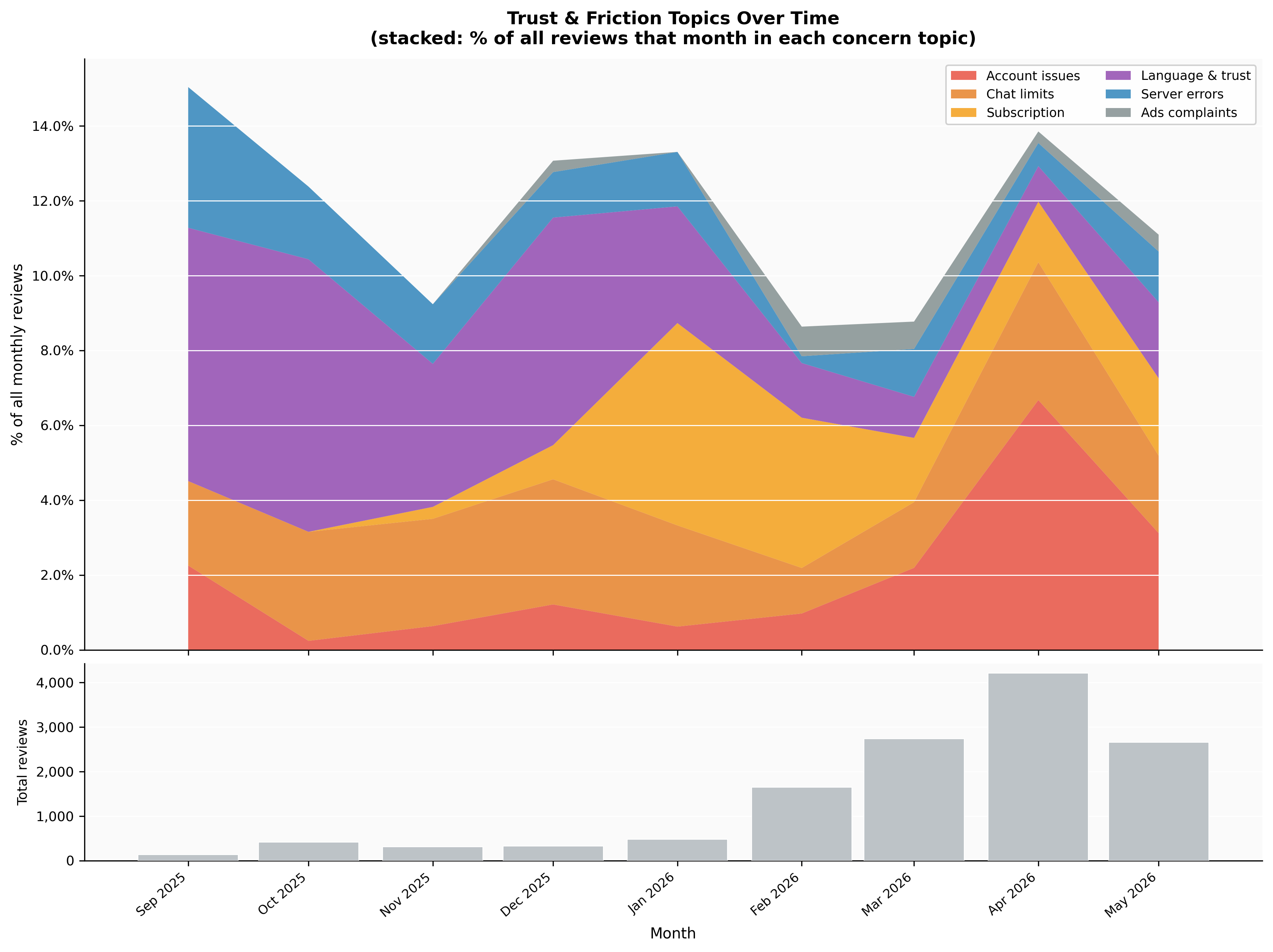}
\caption{Ratio of trust and friction items over time (stacked area chart). Topics related to account issues, subscription complaints, and language/trust are consistently recurring, whereas server problems appear during peak activity months.}
\label{fig:trust-time}
\end{figure*}

As depicted in Figure~\ref{fig:trust-time}, the time-based nature of these trust and friction topics can be observed in terms of the proportion of trust topics (T05, T12, T13, T14). Two prominent topics relating to account and subscription problems were found to persist throughout the data collection period, implying structural rather than sporadic issues.

\begin{figure*}[t]
\centering
\includegraphics[width=\textwidth]{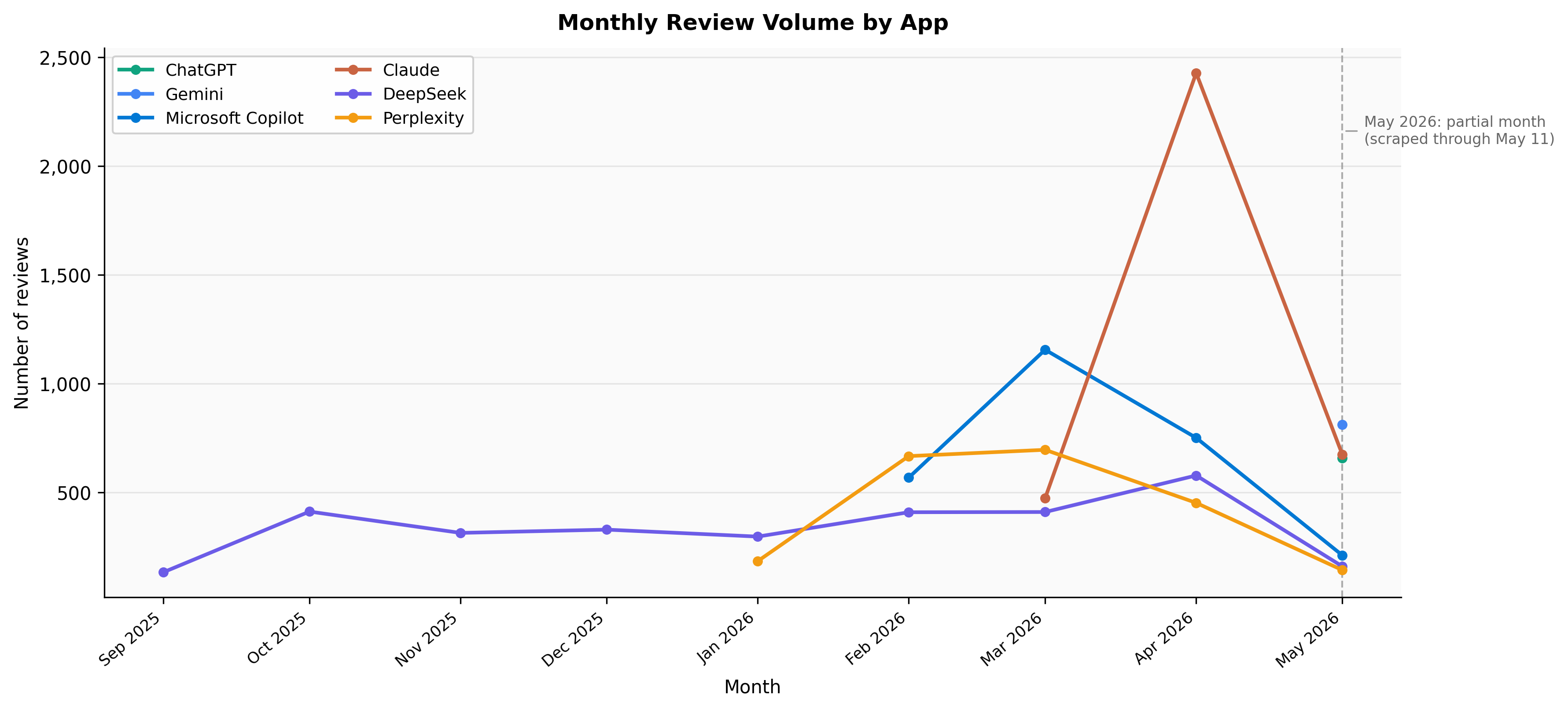}
\caption{Monthly review volume per application. Claude shows a sharp spike in March--April 2026, coinciding with major feature releases. DeepSeek maintains steady volume from its September 2025 launch.}
\label{fig:timeline}
\end{figure*}

Figure~\ref{fig:timeline} presents the monthly review volume per application, providing context for interpreting the temporal patterns. Claude exhibited a sharp spike in review volume during March and April 2026, coinciding with major feature releases and changes to its free-tier usage limits.

\subsection{Thematic Validation Results}
\label{sec:results-validation}

Manual thematic coding was performed independently by two coders on a stratified sample of 300 reviews. Inter-coder reliability between the two human coders was moderate (Cohen's $\kappa$ = 0.544, Krippendorff's $\alpha$ = 0.543, 62.7\% exact agreement), indicating acceptable reliability for a 12-category coding task \cite{landis1977kappa}. Agreement was highest for the \textit{positive} theme (F1 = 0.706), \textit{account} theme (F1 = 0.667), and \textit{limits} theme (F1 = 0.667), and lowest for the \textit{comparison} theme (F1 = 0.357) and abstract themes such as \textit{trust} and \textit{privacy}. Disagreements (n = 112) were resolved through adjudication by the first author, producing gold-standard labels. Using these gold-standard labels, Cohen's Kappa between BERTopic automated assignments and human coding was $\kappa$ = 0.241, indicating fair agreement. The moderate inter-coder $\kappa$ of 0.544 is itself informative for construct validity: concrete, lexically grounded themes were coded reliably, whereas agreement fell for interpretive categories (\textit{comparison}) and especially for the abstract \textit{trust}, \textit{privacy}, and \textit{usability} themes. Findings that rest on these low-agreement themes are therefore interpreted with particular caution throughout the paper (Section~\ref{sec:threats}). Automatic topic-coherence metrics tell the same story: coherence is highest for concrete, keyword-driven topics (T12 Pricing, $C_v = 0.76$; T05 Account, $C_v = 0.73$) and lowest for broad or abstract clusters (T00 General positive, $C_v = 0.42$; T18 Meta-rating, $C_v = 0.41$), with a model mean $C_v$ of 0.556 and $C_{\text{NPMI}}$ of 0.054 (Appendix~\ref{app:coherence}).

\begin{figure*}[t]
\centering
\includegraphics[width=0.8\textwidth]{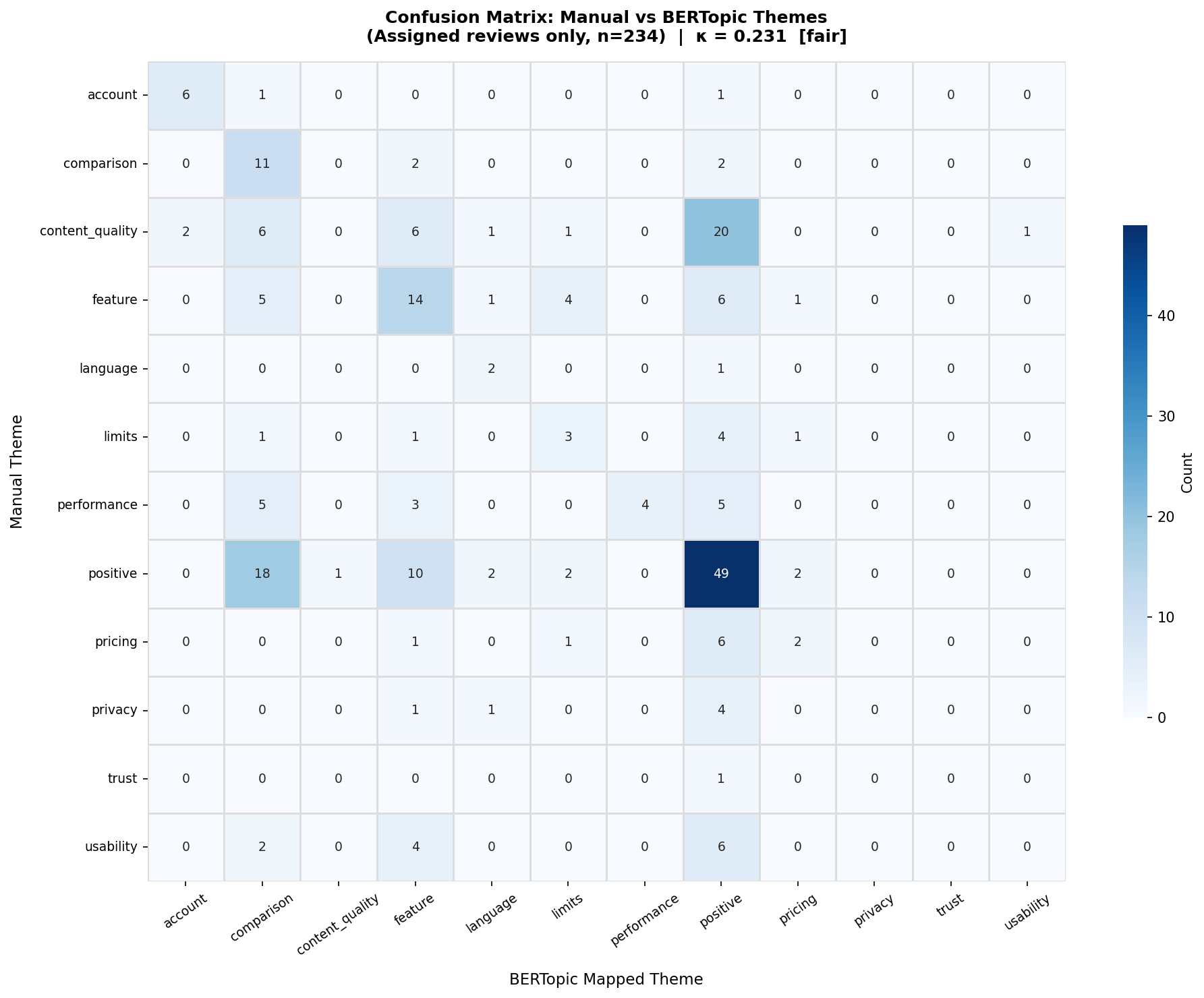}
\caption{Confusion matrix comparing BERTopic automated topic assignments (mapped to manual theme codes) against human thematic coding (n=234 assigned reviews, $\kappa$ = 0.241). BERTopic achieves high precision for lexically distinctive themes (account, comparison) but conflates abstract themes (trust, privacy, usability) into broader positive/feature categories.}
\label{fig:confusion}
\end{figure*}

The confusion matrix (Figure~\ref{fig:confusion}) reveals a systematic pattern of agreement and disagreement. BERTopic had high precision on lexically distinct topics whose keywords were semantically relevant to the contents of the review. The \textit{account} topic (login, phone verification) attained an agreement of 75\%, since the words ``phone,'' ``number,'' ``sign,'' and ``account'' appeared frequently in the reviews assigned to Topic T05, and were also recognized by both automatic and manual classifiers. Likewise, the \textit{comparison} topic and the \textit{language} topic (Chinese/English issues) attained agreements of 73\% and 67\%, respectively.

On the other hand, BERTopic exhibited poor to no performance on abstract topics that do not have clear lexical identifiers. Topics like \textit{trust} ($0\%$ agreement), \textit{privacy} ($0\%$ agreement), and \textit{usability} (close to $0\%$ agreement) were consistently assigned to larger topics by BERTopic, especially T00 (General positive experience) and T01 (AI quality comparisons). This behavior was predictable because trust, privacy, and usability are abstract concepts that people may discuss in various ways without repeatedly mentioning any particular words.

The $\kappa = 0.241$ result in itself represents a methodological contribution. It shows empirically that even though BERTopic represents the most advanced topic modeling technique currently available, it still has inherent limitations when applied to app store reviews related to AI products. This insight highlights the necessity of a complementary manual validation of themes when researchers make an attempt to derive trust and usability-related conclusions based on the results of automated text mining processes.

\begin{figure*}[t]
\centering
\includegraphics[width=0.8\textwidth]{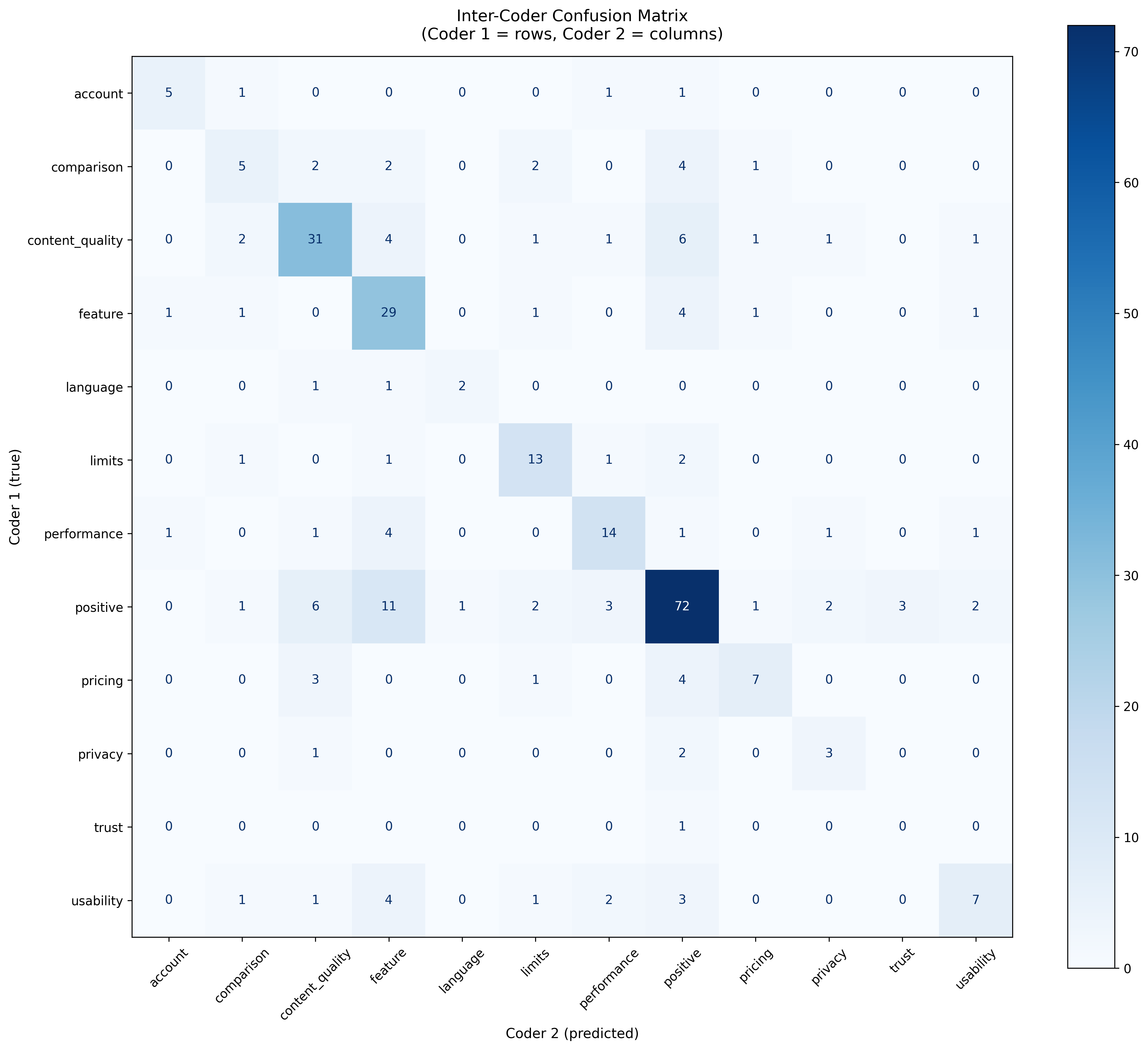}
\caption{Confusion matrix between two independent human coders (n=300). 
Cohen's $\kappa$ = 0.544 (moderate agreement). Strongest agreement on 
\textit{positive} and \textit{content\_quality} themes; weakest on 
\textit{comparison} and \textit{usability}.}
\label{fig:intercoder}
\end{figure*}
\subsection{Sentiment Model Validation}
\label{sec:results-sentiment-validation}

\begin{figure*}[t]
\centering
\includegraphics[width=\columnwidth]{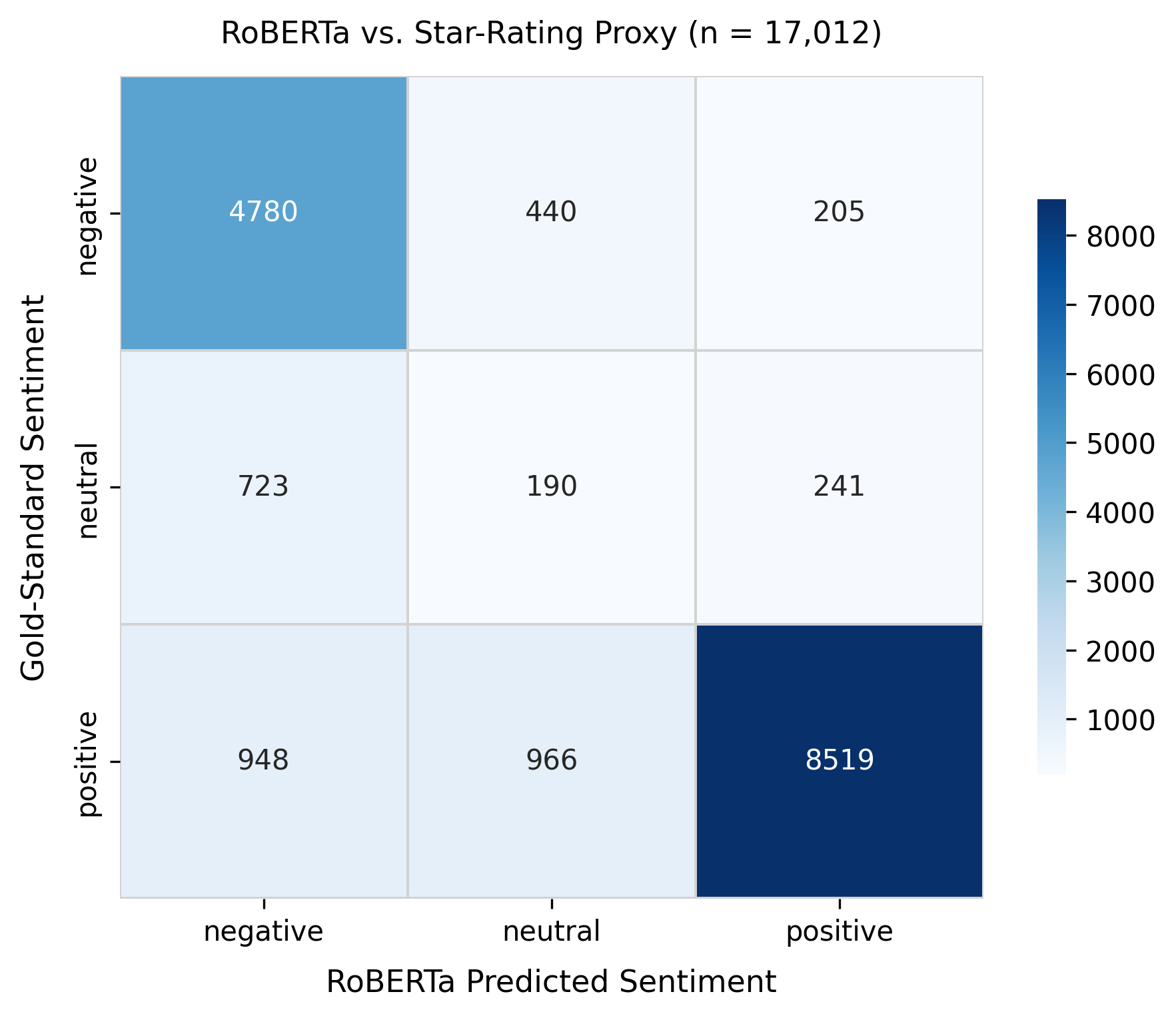}
\caption{RoBERTa sentiment versus the star-rating proxy (accuracy = 79.3\%, macro-$F_1$ = 0.607). Most disagreements involve 3-star reviews.}
\label{fig:sentconfusion}
\end{figure*}

Table~\ref{tab:sentval} reports the validation of the RoBERTa sentiment classifier. In the preliminary star-rating proxy check across all 17,012 reviews, RoBERTa reached an overall accuracy of 79.3\% and a macro-averaged $F_1$ of 0.607. Agreement with the proxy was strong for the \textit{positive} class ($F_1$ = 0.878, precision = 0.950) and the \textit{negative} class ($F_1$ = 0.805, recall = 0.881), but weak for the \textit{neutral} class ($F_1$ = 0.138). The confusion matrix (Figure~\ref{fig:sentconfusion}) shows that most of the error concentrates in the neutral band: reviews rated three stars are frequently written in clearly positive or negative language, so the model---correctly reading the text---diverges from the rating-derived proxy. This pattern is a known property of star-versus-text comparisons and indicates that the neutral proxy, rather than the classifier, is the main source of apparent disagreement.

\begin{figure*}[t]
\centering
\includegraphics[width=\columnwidth]{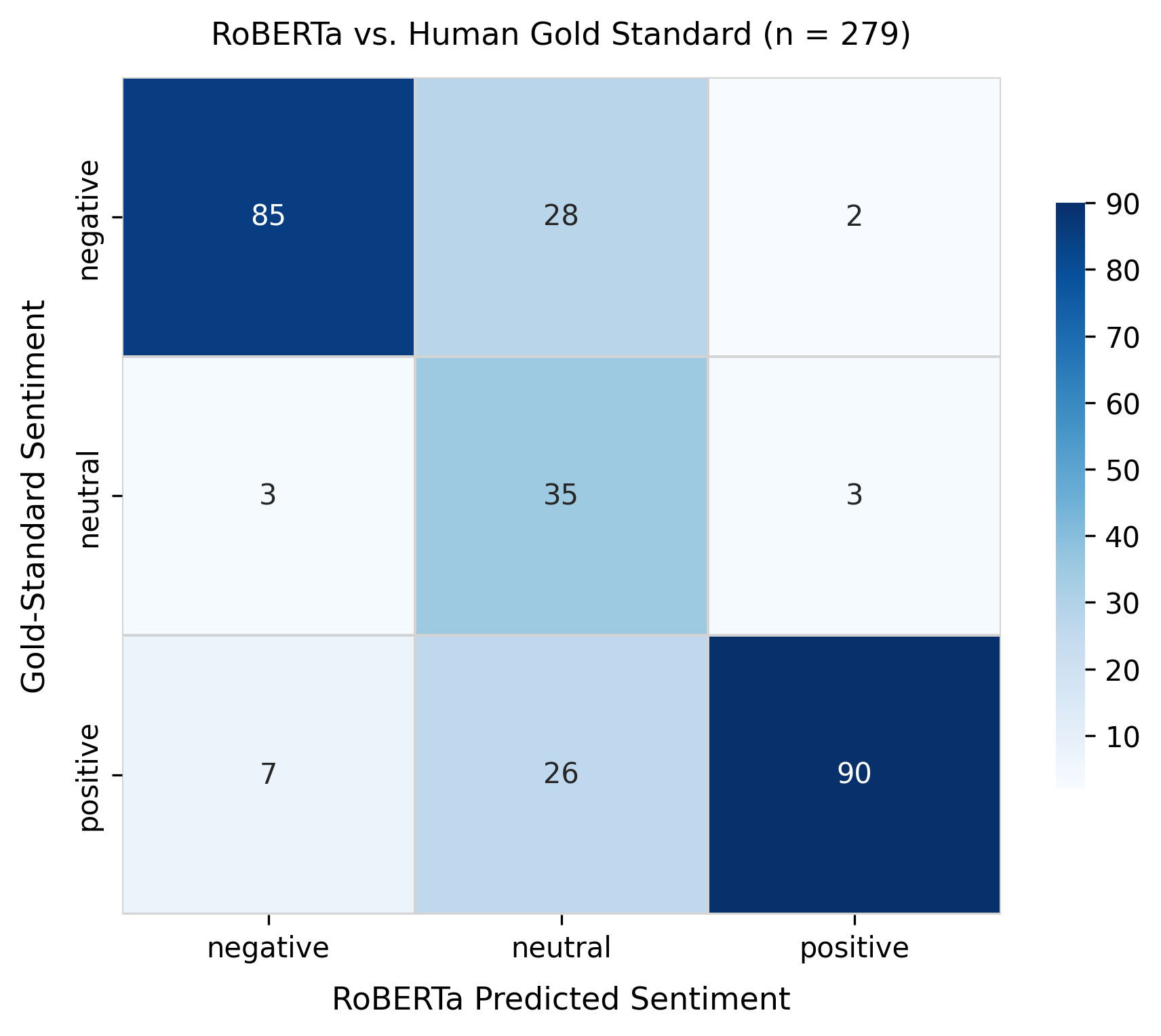}
\caption{RoBERTa sentiment versus human annotations ($n=279$, accuracy = 75.3\%, macro-$F_1$ = 0.725). Errors are concentrated at the neutral boundary.}
\label{fig:sentconfusionhuman}
\end{figure*}

Because the star-rating proxy is only a weak label, we additionally validated RoBERTa against human sentiment coding of the stratified 300-review sample (Section~\ref{sec:method-sentiment-validation}). Two coders independently labelled each review's sentiment from its text alone; 279 reviews received a label from both coders. Inter-coder agreement was very high (Cohen's $\kappa$ = 0.894, Krippendorff's $\alpha$ = 0.894, 93.6\% exact agreement), substantially exceeding the agreement obtained for the more difficult 12-category thematic coding task and indicating that sentiment is a reliably codeable construct. Against the adjudicated human gold standard, RoBERTa achieved an accuracy of 75.3\% and a macro-averaged $F_1$ of 0.725. Performance was strong for the \textit{negative} ($F_1$ = 0.810, precision = 0.895) and \textit{positive} ($F_1$ = 0.826, precision = 0.947) classes and weaker for the \textit{neutral} class ($F_1$ = 0.539): the model assigned the neutral label more liberally than the human coders (89 vs.\ 41 reviews), pulling some mildly-valenced positive and negative reviews into the neutral band. The confusion matrix against human labels (Figure~\ref{fig:sentconfusionhuman}) confirms that errors are concentrated in the neutral boundary rather than in confusions between positive and negative, so the classifier's polarity judgements---which drive the study's substantive findings---are well supported. The high precision on the negative class (0.895) is particularly reassuring given that the paper's conclusions centre on negative-sentiment friction themes.

\begin{table}[t]
\centering
\caption{Validation of the RoBERTa sentiment classifier. The star-rating proxy treats 1--2 stars as negative, 3 as neutral, and 4--5 as positive across the full corpus; the human-coded gold standard is the adjudicated two-coder labelling of the stratified validation sample (279 doubly-labelled reviews).}
\label{tab:sentval}
\small
\begin{tabular}{lrr}
\toprule
Metric & Star-rating proxy & Human-coded \\
       & (n = 17{,}012)    & (n = 279) \\
\midrule
Accuracy            & 0.793 & 0.753 \\
Macro-$F_1$         & 0.607 & 0.725 \\
Weighted-$F_1$      & 0.805 & 0.777 \\
\midrule
Negative $F_1$      & 0.805 & 0.810 \\
Neutral $F_1$       & 0.138 & 0.539 \\
Positive $F_1$      & 0.878 & 0.826 \\
\midrule
Inter-coder $\kappa$      & --- & 0.894 \\
Krippendorff's $\alpha$   & --- & 0.894 \\
\bottomrule
\end{tabular}
\end{table}

\subsection{Multivariate Sentiment Modeling}
\label{sec:results-mlr}
Table~\ref{tab:mlr-main} presents key results from the multinomial logistic regression (Equation~\ref{eq:mlr}). The full model was significant against the intercept-only null ($\chi^2(28) = 3{,}941.1$, $p < .001$) with McFadden's pseudo-$R^2 = 0.125$.

\begin{table*}[t]
\centering
\caption{Multinomial logistic regression: selected odds ratios (OR) for negative sentiment relative to positive. Reference: App = ChatGPT, Topic cluster = Experience \& Sentiment. Only predictors with $p < .05$ shown.}
\label{tab:mlr-main}
\small
\begin{tabular}{llccc}
\toprule
\textbf{Predictor} & \textbf{Category} & \textbf{OR (Neg/Pos)} & \textbf{95\% CI} & \textbf{$p$} \\
\midrule
\multicolumn{5}{l}{\textit{Application (ref: ChatGPT)}} \\
& Claude             & 1.56 & [1.37, 1.79] & $< .001$ \\
& Perplexity         & 1.60 & [1.39, 1.84] & $< .001$ \\
& Gemini             & 1.43 & [1.17, 1.76] & $< .001$ \\
& DeepSeek           & 0.79 & [0.66, 0.94] & .010 \\
& Microsoft Copilot  & 0.81 & [0.69, 0.96] & .013 \\
\midrule
\multicolumn{5}{l}{\textit{Topic cluster (ref: Experience \& Sentiment)}} \\
& Issues (T05, T12, T14, T16)       & 19.06 & [16.04, 22.66] & $< .001$ \\
& Features \& UI (T03, T04, T11)    & 6.10  & [5.22, 7.12]   & $< .001$ \\
& Outliers (T-1)                     & 3.46  & [3.07, 3.89]   & $< .001$ \\
& Specialist (T17--T23)              & 2.92  & [2.23, 3.82]   & $< .001$ \\
& Comparisons (T02, T07)            & 1.55  & [1.37, 1.75]   & $< .001$ \\
& App-specific (T06--T10)           & 0.65  & [0.55, 0.76]   & $< .001$ \\
\midrule
\multicolumn{5}{l}{\textit{Controls}} \\
& Platform (iOS)       & 0.92 & [0.84, 1.01] & .093 (ns) \\
& Log(word count)      & 1.88 & [1.78, 1.98] & $< .001$ \\
& Month                & 0.97 & [0.94, 1.00] & .038 \\
\midrule
\multicolumn{5}{l}{\textit{Model fit}} \\
& McFadden pseudo-$R^2$ & \multicolumn{3}{l}{0.125} \\
& AIC                    & \multicolumn{3}{l}{27,669} \\
& LR $\chi^2$ (vs null)  & \multicolumn{3}{l}{3,941.1 ($p < .001$)} \\
& $N$                    & \multicolumn{3}{l}{17,012} \\
\midrule
\multicolumn{5}{l}{\textit{Interaction model (App $\times$ friction topics)}} \\
& $\Delta$AIC            & \multicolumn{3}{l}{$-45.5$ (improved fit)} \\
& LR $\chi^2$(10)        & \multicolumn{3}{l}{65.5 ($p < .001$)} \\
\bottomrule
\end{tabular}
\end{table*}

\paragraph{Application effects after controlling for topic.}
After controlling for topic membership, Claude's adjusted odds ratio for negative sentiment relative to ChatGPT is 1.56 (95\% CI [1.37, 1.79]), indicating that Claude's elevated negativity is only partially attributable to friction-topic concentration; an inherent app-level effect persists.

\paragraph{Topic effects: quantifying friction severity.}
Holding all other factors constant, the Issues cluster (containing T05, T12, T14, T16) has the strongest association with negative sentiment (OR = 19.06, $p < .001$), followed by Features \& UI (OR = 6.10) and Specialist topics (OR = 2.92). Longer reviews are also more negative (OR = 1.88 per log-unit increase in word count). All the trust-friction themes yield significantly higher odds ratios than the feature/feedback and comparison themes, thus validating the idea that trust and accessibility challenges are more harmful to sentiment than feature limitations.

\paragraph{Interaction effects.}
The extended model with App $\times$ Topic interactions (Equation~\ref{eq:mlr-interaction}) improved fit ($\Delta$AIC = $-45.5$, LR $\chi^2(10) = 65.5$, $p < .001$), indicating that friction topics affect sentiment differently across applications. For example, account friction (T05) has a disproportionately stronger negative effect on Claude than on other applications, consistent with Claude's phone verification requirement generating uniquely intense frustration.

\subsection{Trust Friction Scores}
\label{sec:results-tfs}
Table~\ref{tab:tfs} presents the composite Trust Friction Score and its five sub-dimensional components (Equations~\ref{eq:tfs}--\ref{eq:tfs-ads}).

\begin{table*}[t]
\centering
\caption{Trust Friction Scores (TFS) per application (\%). Higher = greater friction exposure. Highest per column in \textbf{bold}. 95\% bootstrap CIs for composite TFS in brackets.}
\label{tab:tfs}
\small
\begin{tabular}{lccccccc}
\toprule
\textbf{App} & \textbf{TFS$^{\text{auth}}$} & \textbf{TFS$^{\text{limits}}$} & \textbf{TFS$^{\text{price}}$} & \textbf{TFS$^{\text{geo}}$} & \textbf{TFS$^{\text{reliab}}$} & \textbf{TFS$^{\text{ads}}$} & \textbf{TFS (composite)} \\
\midrule
ChatGPT   & 0.38 & 1.20 & 0.60 & 0.16 & 0.38 & 0.11 & 2.83 [2.12, 3.64] \\
Gemini    & 0.25 & 0.74 & 1.11 & 0.99 & \textbf{1.72} & 0.86 & 5.67 [4.19, 7.39] \\
Copilot   & 0.75 & 0.52 & 0.07 & 0.30 & 0.41 & \textbf{1.23} & 3.28 [2.61, 3.98] \\
Claude    & \textbf{8.28} & \textbf{3.26} & 1.71 & 0.14 & 0.62 & 0.04 & \textbf{14.04} [13.06, 15.01] \\
DeepSeek  & 0.89 & 1.02 & 0.03 & \textbf{2.66} & 1.87 & 0.03 & 6.51 [5.65, 7.43] \\
Perplexity& 0.81 & 0.33 & \textbf{4.87} & 0.33 & 0.45 & 0.39 & 7.18 [6.34, 8.01] \\
\bottomrule
\end{tabular}
\\[2pt]\footnotesize{\textit{Note:} Spearman $\rho$(TFS, negative\%) = 0.886, $p$ = .019 ($n = 6$; illustrative, not inferential---see text). Sub-scores: auth = account/sign-in (T05), limits = chat limits (T11), price = subscription (T12), geo = language/trust (T13), reliab = server errors (T14), ads = advertising (T16); the six sub-scores sum to the composite TFS.}
\end{table*}

\paragraph{Composite TFS rankings.}
Claude exhibits the highest composite TFS (14.04\%), driven overwhelmingly by authentication friction ($\text{TFS}^{\text{auth}} = 8.28\%$) and chat limits ($\text{TFS}^{\text{limits}} = 3.26\%$). Perplexity ranks second (TFS = 7.18\%), dominated by pricing friction ($\text{TFS}^{\text{price}} = 4.87\%$). DeepSeek ranks third (TFS = 6.51\%), with a distinctive profile driven by geopolitical trust ($\text{TFS}^{\text{geo}} = 2.66\%$) and server reliability ($\text{TFS}^{\text{reliab}} = 1.87\%$).

\paragraph{Sub-dimensional trust profiles.}
Each application has a distinct friction profile: Claude is dominated by authentication friction, DeepSeek by geopolitical and reliability concerns, Microsoft Copilot by advertising, and Perplexity by pricing. ChatGPT and Gemini exhibit the lowest composite TFS with no dominant friction dimension.

\paragraph{Correlation with overall sentiment.}
Spearman's rank correlation between composite TFS and negative review proportion yields $\rho = 0.886$ ($p = .019$, $n = 6$), consistent with TFS behaving as a summary of trust-related dissatisfaction. This correlation is computed over only six applications and we treat it as \emph{illustrative rather than inferential}: a nonparametric bootstrap over the six apps produces a 95\% confidence interval for $\rho$ that spans essentially the entire admissible range $[0,1]$, so the point estimate must not be read as strong evidence of association. We report it to show internal consistency of the metric, not to establish a population-level relationship.

\paragraph{Comparison with simpler baselines.}
We benchmarked TFS against two scalar baselines: each application's overall negative-sentiment rate, and its friction-topic prevalence (the share of reviews falling in any friction topic, ignoring sentiment). The three measures rank the six applications very similarly (Spearman $\rho$ = 0.886 for TFS vs.\ negativity and for TFS vs.\ prevalence), which is expected because TFS is, by construction, the fraction of an application's reviews that are friction-related negatives. We therefore do not claim that TFS yields a materially different ordering from negative sentiment alone; its contribution is diagnostic rather than ordinal. Unlike a scalar negativity rate, TFS decomposes each application's friction into interpretable sub-dimensions that point to different interventions---Claude's score is dominated by authentication ($\text{TFS}^{\text{auth}}$ = 8.28 of 14.04), Perplexity's by pricing (4.87 of 7.18), DeepSeek's by geopolitical and reliability concerns, and Microsoft Copilot's by advertising---information a single negativity figure cannot convey. The full baseline and robustness results are reported in Appendix~\ref{app:tfsrobust-sec}, Table~\ref{tab:tfsrobust}.

\paragraph{Robustness to weighting and definition.}
We recomputed TFS under nine variants: the main prevalence-weighted definition, an equal-weighting of the six sub-dimensions, a star-rating-based negativity ($\text{score} \le 2$) in place of the RoBERTa label, and six leave-one-topic-out definitions. Claude retained the highest composite TFS in eight of the nine variants, and the ranking was stable when negativity was redefined from star ratings (Spearman $\rho$ = 0.94 vs.\ the main ranking). Two sensitivities are worth stating plainly. First, because Claude's TFS is so heavily driven by authentication, removing the authentication topic moves Perplexity (a pricing-dominated profile) into first place; the claim that Claude is the most friction-exposed application is thus specifically an authentication story. Second, equal-weighting the sub-dimensions (discarding prevalence) reorders the mid-ranked applications ($\rho$ = 0.43 vs.\ the main ranking), confirming that prevalence weighting is a consequential modelling choice and that TFS should be interpreted together with its sub-scores rather than as a single opaque index.

\paragraph{Practical interpretation.}
The TFS model allows for a diagnostic approach: product teams can determine their highest-scoring trust dimension and design interventions accordingly. Measuring sub-scores over time (e.g., $\text{TFS}^{\text{auth}}$ without phone verification) will determine the success of the interventions.

\subsection{Robustness to Sampling Imbalance}
\label{sec:results-robustness}

The six applications contribute unequal numbers of reviews (from 812 for Gemini to 5,037 for Claude), which could in principle let the larger corpora dominate the cross-application comparisons. We therefore assessed the robustness of our key findings to this imbalance in three ways; full results are reported in Appendix~\ref{app:sensitivity}.

First, we repeated the two central omnibus tests on \emph{balanced} subsamples, downsampling every application to the smallest app's size ($n = 812$ per app; 4,872 reviews per draw) across 1,000 independent random draws. The cross-application difference in sentiment remained statistically significant (chi-square) in 100\% of draws, with a scale-free effect size essentially identical to the full sample (Cram\'er's $V$ = 0.122 balanced vs.\ 0.128 full). The difference in star ratings (Kruskal--Wallis) was likewise significant in 100\% of draws ($\eta^2$ = 0.032 balanced vs.\ 0.037 full). The absolute test statistics are smaller under balancing only because $N$ is smaller; the effect sizes, which are the appropriate scale-invariant comparison, are unchanged. The rank ordering of applications by negativity was highly stable: Claude retained the highest negative-sentiment rate in 99.2\% of draws (mean rank 1.01) and Microsoft Copilot the lowest in 99.8\%, with per-application rates matching the full-sample values to within roughly one percentage point.

Second, we stratified by platform (Appendix~\ref{app:sensitivity}, Table~\ref{tab:platform}). Only three applications (ChatGPT, Claude, Perplexity) appear on both the Apple App Store and Google Play; DeepSeek, Gemini, and Microsoft Copilot are Android-only. Where both platforms are available, platform effects are small and application-specific: Claude's elevated negativity is essentially identical across platforms (47.5\% Android vs.\ 48.1\% iOS; $\chi^2 = 1.4$, $p = .49$), whereas ChatGPT is more negative on iOS (35.3\% vs.\ 22.9\%; $V$ = 0.13) and Perplexity more negative on Android (45.3\% vs.\ 37.9\%; $V$ = 0.08). The headline Claude polarization finding therefore does not depend on platform composition.

Third, an equal-weight aggregate---averaging the six per-application negativity rates rather than pooling raw reviews---yields 36.1\%, close to the raw pooled figure of 37.9\%, confirming that corpus-level summaries are not an artifact of Claude's larger share. Taken together, these checks indicate that the study's cross-application conclusions are robust to the sampling imbalance.

\section{Discussion}
\label{sec:discussion}

The current section will be devoted to the interpretation of the most important findings of our research and their relation to the extant literature on trust, usability, and adoption of AI-based systems.

\subsection{Key Findings and Interpretation}
\label{sec:disc-findings}

\paragraph{Authentication friction as a trust-destroying barrier.}
The discovery that problems with signing in and accounts (T05) resulted in 89\% negative sentiment renders authentication friction the second most harmful problem in the dataset, after advertising. Such findings hold great theoretical importance once examined from the perspective of the Technology Acceptance Model (TAM) \cite{davis1989tam}. According to TAM theory, perceived ease of use is one of the principal predictors of acceptance of technology; thus, the fact that authentication problems constitute a powerful impediment to ease of use, occurring right at the beginning of using the tool, before experiencing its functionality, becomes particularly relevant. The trust model proposed by Hoff and Bashir \cite{hoff2015trust} helps shed more light on this phenomenon: the initial level of trust established with respect to a technology is greatly impacted by first impressions and early experiences with the said technology; thus, a negative experience with the sign-in process could potentially harm any learning-based trust that has yet to develop. Claude emerged as the predominant source of literature on this subject matter, possibly due to the necessity for phone number verification, which multiple sources noted as being too stringent when attempting to access an AI chatbot. 

\paragraph{The Claude polarization paradox.}
One of the most fascinating observations is that Claude both has the largest percentage of negative sentiments (47.7\%) and, at the same time, one of the most enthusiastic user groups for the positive sentiment class. In the application-specific topic T08 (Claude-related feedback), the positive sentiment rate for users mentioning Claude was 78\%, whereas the sentiment distribution had a highly bimodal form (32.3\% one-star, 40.2\% five-star). This observation indicates that there might be two separate populations of users for Claude: the first one consists of people interested in its technical aspects and valuing its qualities (such as reasoning capability, safety, and conversation style), whereas the second one encounters barriers (such as authentication, message limitations, and subscription requirements) to use those features. This analysis is aligned with the diffusion of innovation framework, whereby early adopters are expected to judge products based on their capabilities, whereas the early majority considers ease of access and usefulness \cite{rogers2003diffusion}. In terms of responsible adoption, the example of Claude demonstrates how structural factors might result in a seemingly poor product from the standpoint of mainstream consumers despite positive reception within the core user base.
\paragraph{Quantifying the polarization paradox.}
The formal polarization analysis (Table~\ref{tab:polarization}) transforms the Claude paradox from a qualitative observation into a measured phenomenon. Interestingly, ChatGPT exhibits the highest bimodality coefficient ($BC = 0.896$) while Perplexity achieves the highest Esteban-Ray index ($ER_{1.0} = 0.652$). With a BC of 0.814, Claude's is actually the lowest, implying that its polarization is reflected more in the intensity of sentiments rather than the star-rating extremes. Alongside the finding from the multinomial regression analysis, which indicates that Claude's negativity is only partially accounted for by the friction topic composition (OR = 1.56 when adjusted for topics, Section~\ref{sec:results-mlr}), this result implies that the poles represent (a) technologically savvy individuals who circumvent friction and assess model performance and (b) average users where friction is the overwhelming experience. The TFS analysis (Table~\ref{tab:tfs}) identifies $\text{TFS}^{\text{auth}}$ as the major contributing factor, implying that lowering authentication friction would move masses from the 1-star pole to the center without affecting the 5-star pole.

\paragraph{Geopolitical trust as an underexplored dimension.}
Finding Topic T13 (Language and trust, Chinese; $n = 253$, 47\% negative) is, to the best of our knowledge, among the first observations of such a topic in the app store review mining literature. The topic highlighted two important issues that were specific to the DeepSeek application. They included an issue relating to the functional usability of the application (that is, it would respond in Chinese when prompted to use English commands), and the second issue revolved around the users' mistrust of the application because it originated from China. This was in light of geopolitical tension surrounding the DeepSeek application that was reported in the security literature, including the banning by governments and the presence of safety threats, as well as the existence of user data in Chinese servers \cite{deepseek2025security}. From a conceptual perspective, this study contributes to existing theories on trust by suggesting that trust in artificial intelligence technologies may not be merely a function of competence, benevolence, and integrity \cite{mayer1995trust} but could also be influenced by geopolitical considerations related to the nature of the provider. The framework proposed by Siau and Wang \cite{siau2018building} identifies culture as a variable influencing trust in AI technologies, but the analysis of the DeepSeek case indicates that the nationalities of the providers and corresponding governing structures can themselves operate as a separate trust dimension, which cannot be easily overcome by technology enhancements alone. Because topic modeling captures such abstract constructs only weakly (Section~\ref{sec:threats}), we advance this as an exploratory, theory-generating observation drawn from the language of a subset of reviews rather than as a confirmed or quantified effect.

\paragraph{Subscription pricing as an adoption barrier.}
T12 (Pricing and Subscription; 73\% negative) indicates that the freemium business model adopted by most of the generative AI products is causing significant user dissatisfaction. The participants from different products, with Perplexity and Claude making up the largest share of mentions, showed their displeasure about artificial restrictions imposed on the free version, sudden paywalls for services they deemed as basic, and a lack of clarity about what the paid version provides compared to the free one. This insight relates to the research on digital product pricing and fairness perception \cite{venkatesh2012consumer}. In terms of responsible adoption, unclear pricing strategies can serve as an obstacle for equal opportunity, particularly for users operating in low socio-economic environments unable to test out the paid version to evaluate its value proposition.

\paragraph{Advertising as a rejection signal.}
The near-universal negativity of Topic T16 (Ads; 91\% negative), concentrated in Microsoft Copilot reviews, suggests that users hold generative AI applications to a different standard than other mobile applications when it comes to advertising. While advertising is a common and generally tolerated monetization strategy in mobile apps, our data indicate that users perceive ads within an AI assistant as fundamentally inappropriate, likely because the conversational, trust-dependent nature of AI interaction is incompatible with the attention-diverting and commercially motivated nature of advertising.

\paragraph{Comparison with prior work.}
Our findings extend and partially diverge from the two existing GenAI app review studies. Alabduljabbar \cite{alabduljabbar2024} found that ChatGPT achieved the highest compound usability scores among the five applications studied, a finding broadly consistent with our data showing ChatGPT's relatively high positive sentiment (60.6\%) and average rating (3.86). However, our BERTopic-based analysis reveals thematic nuances that Alabduljabbar's VADER-plus-LDA pipeline could not detect, including the distinct authentication friction, pricing, and geopolitical trust themes that emerge only with contextual embeddings. Meng et al. \cite{meng2026} extracted feature-related topics from their dataset of 100,000 reviews; however, since Meng et al.'s analysis did not incorporate a framework related to trust or adoption, no distinction was made between friction topics such as signing-in problems and subscription obstacles, as opposed to feature-related topics. Our research shows how analyzing reviews using a trust and usability perspective generates entirely new insights.

\subsection{Implications for Research}
\label{sec:disc-research}

The paper makes a contribution to the literature based on five counts. Firstly, the analysis shows that the review mining approach that is very common in software engineering when researching traditional apps \cite{martin2017survey} could also be applied to generative AI, providing a new avenue for investigating questions related to trust, usability, and acceptance of technology beyond surveys and experiments. Secondly, the naturalistic quality of the reviews allows the researcher to explore issues of trust (geopolitical), authentication, and pricing,g which may not come up in controlled experiments where the participants get access to the system for free.

Second, the use of BERTopic in conjunction with \\ transformer-based sentiment classification constitutes an innovation from the methodology used in previous research on GenAI reviews, which utilized LDA followed by VADER \cite{alabduljabbar2024}. The contextual embeddings generated by BERTopic provide greater coherence in topics, while the sentiment classifier based on RoBERTa allows for better capture of contextual sentiment (considering negation or partial praise). Future researchers conducting app store reviews would do well to employ this pipeline as their default option, considering the relatively low computational cost compared to LDA.

Thirdly, the result of the Cohen’s Kappa test ($\kappa = 0.241$) is a methodological result on its own. The results clearly indicate that there are inherent limitations to the use of automated topic modeling approaches when applied to more abstract concepts like trust, privacy, and usability. Thus, the study suggests a need for the use of manual thematic coding alongside automated topic models for research in which theory is drawn upon \cite{braun2006thematic}. It means that future work in the area cannot view automated topic modeling as a tool for the replacement of manual thematic coding, but as one of the discovery tools instead.

Fourth, we propose the Trust Friction Score (TFS), an index that represents a multi-dimensional measure of trust that emerges out of our investigation into the topic. While negativity is correlated with both the prevalence of negative topics and their negativity strength, TFS takes into account both of these aspects simultaneously in a unified measure. The breakdown to its constituent dimensions (Equations~\ref{eq:tfs}--\ref{eq:tfs-ads}) offers a generalizable solution to measuring trust friction in ASR research.

Fifth, our finding that trust in generative AI applications operates along multiple independent dimensions, including performance trust, authentication trust, pricing trust, and geopolitical trust, suggests that existing unidimensional or even bidimensional trust scales may be insufficient for capturing the full range of user concerns in this domain \cite{mayer1995trust, hoff2015trust}. We encourage researchers to develop and validate multi-dimensional trust measurement instruments specifically calibrated for consumer-facing generative AI applications.

\subsection{Implications for Practice}
\label{sec:disc-practice}

Our findings yield five actionable recommendations for developers and product teams building consumer-facing generative AI applications.

First, \textit{streamline authentication flows as the highest-priority trust intervention}. The 89\% negativity rate for sign-in and account issues indicates that authentication friction is the single most impactful quick win available to developers. Requiring phone number verification for an AI chatbot appears disproportionate to users and should be reconsidered in favor of lighter authentication mechanisms. Amershi et al.'s \cite{amershi2019guidelines} guidelines for human-AI interaction recommend making clear what the system can do at the outset of interaction; this principle should be extended to ensuring that users can reach the system without unnecessary barriers.

Second, \textit{implement transparent and well-articulated pricing policies}. Considering the negative attitude towards subscription and pricing (73\%), it is apparent that users do not trust the developers due to the lack of transparency associated with the free tier. It is important that developers offer comparisons between the free and premium versions when users are prompted to upgrade their accounts, and that any form of restriction placed on the essential functions does not feel like punishment to the user \cite{weisz2024design}.

Third, \textit{focus on language and localization quality when designing applications targeting global customers}. The mistrust that exists among users of the DeepSeek application in Chinese is partly because of a failure in localization, where the application does not respond in the user’s language. However, it is also partly a matter of mistrust.

The fourth suggestion is to \textit{not advertise through conversational AI platforms}. The fact that Microsoft's Copilot received an approval rating of only 9\% for its advertisements clearly shows user disapproval. The very essence of communication via AI cannot be reconciled with advertisements, as users have certain expectations in such contexts \cite{fogg2003prominence}.

Fifth, \textit{communicate the constraints of the AI system truthfully and in advance}. The fact that we observed multiple constraints discussed by users in our data, such as constraints related to the message size limit (T11) or constraints related to server dependability (T14), suggests that users are disappointed not only by the constraints themselves but also by the absence of transparency regarding why they are there. In this context, it is helpful for users to know what the reason behind the constraint is \cite{amershi2019guidelines}.

\section{Threats to Validity}
\label{sec:threats}

We organize threats to validity into four categories following established guidelines for empirical software engineering research \cite{wohlin2012experimentation}.

\paragraph{Internal validity.}
Two methodological choices may affect the internal validity of our findings. First, the sentiment classifier (cardiffnlp/twitter-roberta-base-sentiment-latest) was trained on Twitter data rather than app store reviews \cite{loureiro2022timelms}. App store reviews tend to be longer, more structured, and less colloquial than tweets, which may introduce classification errors at the margins, particularly for reviews with mixed or nuanced sentiment. Although this was partially addressed via thematic validation manually, a sentiment analysis model trained specifically for app store data could increase classification accuracy. Additionally, several BERTopic parameters, including \texttt{min\_cluster\_size=30}, \texttt{n\_components=5}, and \texttt{nr\_topics=25,} were determined experimentally rather than being optimized. With another combination of parameters, there might be a different number of topics formed with varying boundaries, resulting in a completely different thematic map from what we discovered. This was partially addressed using topic validation manually and selecting parameters to maximize interpretability and coherence of topics.

\paragraph{External validity.}
The generalizability of our research results is limited in several ways. First, we have examined only six AI generators; other popular platforms like Grok, Pi, Poe, and Character.ai might provoke distinct user apprehensions. Secondly, our analysis was conducted based on reviews in English, which does not include viewpoints from leading consumer markets like India, Brazil, Japan, and countries with Chinese speakers. Since one of our principal results touches upon geopolitical trust, this language limitation can lead us to underestimate the importance and strength of cross-cultural trust interactions. Furthermore, our dataset is unbalanced across applications: Claude provided 5,037 reviews (29.6\% of our entire dataset), whereas Gemini offered only 812 reviews (4.8\%). To ensure this imbalance does not drive our conclusions, we conducted balanced-subsample, platform-stratified, and equal-weight sensitivity analyses (Section~\ref{sec:results-robustness}, Appendix~\ref{app:sensitivity}); the cross-application findings held in essentially all balanced draws, so the imbalance affects the precision of individual per-app estimates more than the substantive conclusions. Finally, only three out of six applications had reviews in the Apple App Store, so the platform-stratified comparisons are limited to those applications.

\paragraph{Construct validity.}
The Cohen's Kappa of $\kappa = 0.241$ between automated and manual topic assignments indicates only fair agreement \cite{landis1977kappa}, reflecting BERTopic's inability to reliably capture abstract themes such as trust, privacy, and usability. This means that the topic-level findings for these abstract constructs should be interpreted as conservative lower bounds rather than precise estimates of their prevalence in the corpus. Manual thematic validation was performed by two independent coders on a stratified sample of 300 reviews. Inter-coder reliability reached moderate agreement (Cohen's $\kappa$ = 0.544, Krippendorff's $\alpha$ = 0.543, 62.7\% exact agreement), which is acceptable for a 12-category qualitative coding task. Disagreements (n = 112) were resolved through adjudication by the first author to produce gold-standard labels. While the primary purpose of the validation was diagnostic (identifying where BERTopic succeeds and fails) rather than establishing a gold-standard human coding, a multi-coder design with formal inter-coder reliability assessment would strengthen the construct validity of the thematic analysis \cite{mcdonald2019reliability}.

\paragraph{Interpreting abstract-construct topics.}
A direct consequence of the preceding point is that our trust- and privacy-related findings must be read as exploratory. BERTopic reliably recovers lexically distinctive topics (e.g., account or pricing complaints, where shared keywords make the cluster coherent), but abstract constructs such as trust, privacy, and usability are expressed in heterogeneous language and were largely absorbed into broad positive/feature clusters by the model, yielding near-zero agreement with human coders for these themes (Section~\ref{sec:results-validation}). We therefore do not interpret the size of a trust- or privacy-labelled topic (notably T13) as a measurement of how many users distrust an application. Where we report quantitative figures for these topics, they are \emph{sentiment} rates computed over the reviews in the topic---and the sentiment classifier producing them is validated against human labels (accuracy 75.3\%, macro-$F_1$ = 0.725; Section~\ref{sec:results-sentiment-validation})---rather than prevalence estimates for the abstract construct itself. Claims about trust, privacy, and geopolitical concern throughout the paper are consequently framed as indicative patterns that warrant targeted follow-up (e.g., survey or interview studies) rather than as confirmatory prevalence measures.

\paragraph{Reliability.}
Two data-related decisions affect the reproducibility and completeness of our findings. First, the 10-word minimum length filter removed 34,530 of the 52,080 post-language-filter reviews (66\%), discarding feedback that may be terse but meaningful (e.g., ``crashes constantly'' or ``love it''). We examined the direction of this bias by comparing removed and retained reviews (Appendix~\ref{app:filter}, Table~\ref{tab:filter}). The removed reviews are markedly more positive than the retained ones (mean rating 4.41 vs.\ 3.54 stars; 76.6\% five-star vs.\ 50.8\%), and they over-represent applications whose users tend to leave brief praise (ChatGPT and Gemini), while Claude---whose users write longer, more critical reviews---is comparatively under-represented among the discarded set. The practical consequence is that our corpus is skewed toward longer, more critical reviews, so the absolute negative-sentiment percentages we report are best read as upper bounds relative to the full reviewer population; the cross-application \emph{comparisons}, however, are computed within this same filtered frame and are not undermined by the filter. Although such filtering is essential for effective topic modeling (short reviews lack the context BERTopic embeddings require), a lower threshold would retain more terse feedback at the cost of noisier topics. Secondly, our data was gathered from a single scraping run conducted in May 2026 and therefore reflects a one-time snapshot of a roughly eight-month period (September 2025 to May 2026) rather than a longitudinal study. Consumer opinions can change over time depending on application improvements or changes to the pricing model, and a cross-sectional approach cannot capture that information. Moreover, the constraints of app store scraping APIs can limit the availability of historical data on consumer opinions for any particular application.
\section{Conclusion}
\label{sec:conclusion}

This research aimed to investigate the true feelings, perceptions, and challenges faced by users of AI-based applications by analyzing 17,012 app store reviews for six popular applications, namely ChatGPT, Gemini, Microsoft Copilot, Claude, DeepSeek, and Perplexity. BERTopic topic modeling, RoBERTa-based sentiment analysis, and statistical analysis were used to identify the prominent user issues, sentiment variance between different applications, and the particular trust and usability-related factors most predictive of dissatisfaction.

Our study identified 24 topics, with five being found to be important trust and usability barriers. The biggest usability challenge identified by our study is related to the friction involved with authentication procedures (89\% negative). This suggests that the sign-in procedure is a key trust-busting factor. Advertisements within conversational AI have also been rejected with great intensity (91\% negative) and were specific to Microsoft Copilot. Pricing issues have been flagged as an adoption issue (73\% negative).

Server problems (83\% negative) were seen to be associated with the teething problems that were occurring during the scaling up of AI infrastructure, specifically DeepSeek. Notably, a subset of DeepSeek reviews surfaced an exploratory geopolitical-trust theme (Topic T13), with users raising concerns over the app's Chinese origin, data privacy, and occasional Chinese-language replies; consistent with topic modeling's limited reliability for abstract constructs, we report this as an indicative pattern rather than a prevalence estimate. Additionally, we found that there was a polarization problem in relation to the application Claude, whereby the application recorded the highest rate of negative feedback (47.7\%) yet still had an enthusiastic group of positive users.

This paper’s contributions can be enumerated into three categories. Firstly, this paper offers, to the best of our knowledge, one of the first comparative cross-application studies of generative AI app reviews, with the inclusion of six apps, including three (Claude, DeepSeek, Perplexity) which, to our knowledge, have not been analyzed in prior app store mining studies. Secondly, this paper shows an advancement in methodology where the use of BERTopic with contextual embeddings and RoBERTa sentiment analysis provides much higher thematic granularity compared to the pipeline involving the use of LDA and VADER in previous literature, while simultaneously providing empirical insights into the limitations of automated topic modeling with respect to abstract concepts ($\kappa = 0.241$). Thirdly, this paper presents new empirical insights into issues of trust, usability, and adoption challenges that will guide the research on human factors in generative AI and the practical development of applications.

There are multiple avenues for future research emerging from this paper. For one thing, conducting a longitudinal study would be helpful in identifying whether the friction barriers that have been mentioned here are structural problems or are temporary pains that eventually fade out as the applications evolve. For another thing, expanding our analysis to other languages besides English, such as Hindi, Spanish, Portuguese, and Chinese, would allow us to consider the points of view of large portions of users that have not been considered in the current study and would enable us to analyze the role of trust across cultures better. Fourth, increasing the sample size by including other software applications like Grok, Pi, Poe, and Character.ai, apart from other non-mobile platforms like web interfaces, IDEs, and desktop applications, will increase the generalizability of the findings that we have discovered thus far. Lastly, creating an AI tool-specific sentiment classifier through fine-tuning of existing models will help overcome the limitations associated with tweets that we faced when conducting our study.

\appendix

\section{Sampling-Imbalance Sensitivity Analyses}
\label{app:sensitivity}

This appendix reports the sensitivity analyses summarised in Section~\ref{sec:results-robustness}. Table~\ref{tab:balanced} compares each key cross-application test on the full corpus against balanced subsamples in which every application is downsampled to the smallest app's size ($n = 812$ per application, 4,872 reviews per draw) over 1,000 random draws. Because the balanced $N$ is smaller than the full corpus, the raw test statistics are necessarily lower; the scale-free effect sizes (Cram\'er's $V$, $\eta^2$) are the appropriate basis for comparison and are essentially unchanged. Table~\ref{tab:platform} reports the platform-stratified (iOS vs.\ Android) negativity analysis.

\begin{table}[h]
\centering
\caption{Balanced-subsample sensitivity (1,000 draws, $n = 812$/app). Balanced values are mean [2.5th, 97.5th percentile] across draws. Effect sizes are essentially unchanged from the full sample, and the ordering of applications by negativity is highly stable.}
\label{tab:balanced}
\small
\begin{tabular}{lll}
\toprule
Quantity & Full sample & Balanced subsamples \\
\midrule
Sentiment $\times$ app: $\chi^2$    & 561.3 & 145.8 [110.5, 185.2] \\
\quad Cram\'er's $V$                 & 0.128 & 0.122 [0.107, 0.138] \\
\quad significant ($p<.05$)          & yes   & 100\% of draws \\
Ratings $\times$ app: $H$            & 639.2 & 159.7 [120.0, 205.0] \\
\quad $\eta^2$                       & 0.037 & 0.032 [0.024, 0.041] \\
\quad significant ($p<.05$)          & yes   & 100\% of draws \\
\midrule
Claude highest negativity            & yes   & 99.2\% of draws \\
Copilot lowest negativity            & yes   & 99.8\% of draws \\
\midrule
\multicolumn{3}{l}{\textit{Per-app negative-sentiment rate (\%), balanced mean}} \\
\quad Claude                         & 47.7  & 47.7 \\
\quad Perplexity                     & 42.3  & 42.3 \\
\quad Gemini                         & 38.7  & 38.7 \\
\quad DeepSeek                       & 31.8  & 31.8 \\
\quad ChatGPT                        & 30.9  & 30.9 \\
\quad Microsoft Copilot              & 25.3  & 25.3 \\
\bottomrule
\end{tabular}
\end{table}

\begin{table}[t]
\centering
\caption{Platform-stratified negative-sentiment rates. Only ChatGPT, Claude, and Perplexity appear on both stores; DeepSeek, Gemini, and Microsoft Copilot are Android-only. Within-app tests are chi-square on the sentiment $\times$ platform table.}
\label{tab:platform}
\footnotesize
\setlength{\tabcolsep}{3pt}
\begin{tabular}{lccp{3.2cm}}
\toprule
\textbf{Scope} &
\textbf{Android} &
\textbf{iOS} &
\textbf{Sentiment $\times$ Platform} \\
\midrule
All apps
& 37.0 (12,917)
& 40.8 (4,095)
& --- \\

ChatGPT
& 22.9 (659)
& 35.3 (1,182)
& $\chi^2=30.5$, $p<.001$, $V=0.13$ \\

Claude
& 47.5 (3,576)
& 48.1 (1,461)
& $\chi^2=1.4$, $p=.49$ (ns) \\

Perplexity
& 45.3 (2,142)
& 37.9 (1,452)
& $\chi^2=21.7$, $p<.001$, $V=0.08$ \\
\bottomrule
\end{tabular}
\end{table}

An equal-weight aggregate (mean of the six per-application negativity rates) is 36.1\%, close to the raw pooled 37.9\%, further indicating that corpus-level summaries are not driven by the larger corpora.

\section{Trust Friction Score: Baselines and Robustness}
\label{app:tfsrobust-sec}

Table~\ref{tab:tfsrobust} accompanies the TFS validation in Section~\ref{sec:results-tfs}. It reports how the composite TFS ranking of the six applications behaves under alternative weightings and definitions, alongside its agreement with the two scalar baselines (negative-sentiment rate and friction-topic prevalence). Agreement is measured by Spearman rank correlation against the main prevalence-weighted TFS and against the negativity baseline; the ``top app'' column records which application receives the highest score under each variant.

\begin{table}[t]
\caption{Robustness of the Trust Friction Score (TFS).}
\label{tab:tfsrobust}
\centering
\footnotesize
\begin{tabularx}{\columnwidth}{Xccc}
\toprule
\textbf{TFS definition} &
\textbf{Top app} &
$\boldsymbol{\rho}$ \textbf{vs. TFS} &
$\boldsymbol{\rho}$ \textbf{vs. Neg.} \\
\midrule
Main (prevalence-weighted) & Claude & 1.00 & 0.89 \\
Equal-weight sub-dimensions & Claude & 0.43 & 0.60 \\
Star-rating negativity ($\le2$) & Claude & 0.94 & 0.94 \\
Drop authentication (T05) & Perplexity & 0.94 & 0.83 \\
Drop chat limits (T11) & Claude & 1.00 & 0.89 \\
Drop pricing (T12) & Claude & 0.66 & 0.49 \\
Drop geopolitical (T13) & Claude & 0.94 & 0.94 \\
Drop reliability (T14) & Claude & 1.00 & 0.89 \\
Drop advertising (T16) & Claude & 0.94 & 0.94 \\
\bottomrule
\end{tabularx}

\vspace{2mm}
\footnotesize
\textit{Baseline rank agreement:}
$\rho(\mathrm{TFS},\mathrm{Neg.})=0.89$,
$\rho(\mathrm{TFS},\mathrm{Prev.})=0.89$,
$\rho(\mathrm{Neg.},\mathrm{Prev.})=0.83$.
\end{table}
\section{Minimum-Length Filter: Removed vs.\ Retained Reviews}
\label{app:filter}

Table~\ref{tab:filter} supports the bias analysis in Section~\ref{sec:threats}. It compares the 34,530 reviews removed by the 10-word minimum-length filter against the 17,550 retained (before deduplication), across star rating and application. Removed reviews are substantially more positive and are concentrated in applications whose users tend to leave short praise.

\begin{table}[t]
\caption{Effect of the 10-word minimum review-length filter.}
\label{tab:filter}
\centering
\footnotesize
\begin{tabular}{lrr}
\toprule
 & \textbf{Removed} & \textbf{Retained} \\
 & \textbf{(<10)} & \textbf{($\ge$10)} \\
\midrule
Number of reviews & 34,530 & 17,550 \\
Mean star rating  & 4.41 & 3.54 \\
5-star (\%)       & 76.6 & 50.8 \\
1-star (\%)       & 9.7 & 25.9 \\
\midrule
\multicolumn{3}{l}{\textit{Application share (\%)}}\\
ChatGPT            & 22.5 & 11.2 \\
Gemini             & 18.8 & 5.0 \\
Perplexity         & 17.9 & 21.1 \\
Microsoft Copilot  & 14.5 & 15.6 \\
Claude             & 13.7 & 29.3 \\
DeepSeek           & 12.5 & 17.8 \\
\bottomrule
\end{tabular}

\vspace{2mm}
\footnotesize
\textit{Note:} Results are based on the post-language-filter corpus
($n=52{,}080$). Short reviews (<10 words) are substantially more
positive, with higher mean ratings and a larger proportion of 5-star
reviews. Rating distributions differ significantly between groups
(Mann--Whitney $U$, $p<.001$), indicating that the retained corpus is
biased toward longer, more critical reviews.
\end{table}

\section{Topic Coherence}
\label{app:coherence}

To complement the manual thematic validation (Section~\ref{sec:results-validation}), we computed two standard automatic topic-coherence metrics over the top-10 words of each of the 24 non-outlier topics, using the analysed corpus as reference: the $C_v$ measure and normalised pointwise mutual information ($C_{\text{NPMI}}$). The model attained a mean $C_v$ of 0.556 (median 0.546) and a mean $C_{\text{NPMI}}$ of 0.054 (median 0.045), values typical of BERTopic on noisy short-text review corpora. Consistent with the manual validation, the most coherent topics are lexically distinctive (T12 Subscription/Pricing, $C_v = 0.76$; T05 Sign-in/Account, $C_v = 0.73$), whereas the least coherent are broad, heterogeneous clusters (T00 General positive, $C_v = 0.42$; T18 Meta-rating, $C_v = 0.41$). This pattern corroborates the finding that BERTopic recovers concrete, keyword-driven topics well but abstract or catch-all clusters less reliably.

\bibliographystyle{elsarticle-num}
\bibliography{cas-refs}

\end{document}